\documentclass{article}
\PassOptionsToPackage{dvipsnames,table}{xcolor}
\usepackage{iclr2027_conference,times}

\usepackage{amsmath,amsfonts,bm}

\def\eqref#1{equation~\ref{#1}}

\def\1{\bm{1}}

\DeclareMathAlphabet{\mathsfit}{\encodingdefault}{\sfdefault}{m}{sl}
\SetMathAlphabet{\mathsfit}{bold}{\encodingdefault}{\sfdefault}{bx}{n}

\usepackage{url}
\usepackage{booktabs}
\usepackage{graphicx}
\usepackage{xcolor}
\definecolor{linkblue}{rgb}{0.0,0.2,0.55}
\usepackage[colorlinks=true,allcolors=linkblue]{hyperref}
\usepackage{multirow}
\usepackage{lipsum}
\usepackage{pifont}
\usepackage{mathtools}
\usepackage{subcaption}
\usepackage{wrapfig}
\usepackage[most]{tcolorbox}
\usepackage{makecell}
\usepackage{hhline}
\usepackage{placeins}
\usepackage{fontawesome5}
\newcommand{\ours}{MoOP}
\newcommand{\cmark}{\ding{51}}
\newcommand{\xmark}{\ding{55}}
\newcommand{\graycell}{\cellcolor{gray!15}}

\newcommand{\projecturl}{https://cvlab-kaist.github.io/MoOP}
\newcommand{\codeurl}{https://github.com/cvlab-kaist/MoOP}

\title{Pruning Autonomous Driving VLAs: \\What to Preserve for Reasoning and Action}

\author{Seongchan Kim$^{1}$, Soowon Son$^{1}$, Sangoh Kim$^{1}$, Hyunah Ko$^{1}$, \\
\bf Eunbeen Hong$^{1}$, Hyunkoo Lee$^{1}$, Sanghyun Lee$^{1}$, WonJun Moon$^{1}$, \\
\bf Kihong Kim$^{2}$, Dongyeon Son$^{2}$, Woongchan Nam$^{2}$, Seungryong Kim$^{1,\dagger}$ \\ [6pt]
$^{1}$KAIST \quad $^{2}$LG Innotek \\[6pt]
\href{\projecturl}{\faGlobe\ Project Page} \qquad \href{\codeurl}{\faGithub\ Code}}

\newtcolorbox{finding}{
  colback=gray!10,
  colframe=black,
  boxrule=0.7pt
}

\iclrfinalcopy
\begin{document}

\maketitle

\begingroup
\renewcommand\thefootnote{$\dagger$}
\footnotetext{Corresponding author.}
\endgroup

\lhead{Preprint.}

\begin{abstract}
Reasoning vision--language--action (VLA) models for autonomous driving use chain-of-thought (CoT) reasoning to guide action generation, but incur substantial memory and computational costs.
We study structured pruning for these models, focusing on preserving both scene understanding and trajectory prediction.
Our analysis of Taylor importance in Alpamayo-1.5-10B shows that reasoning and trajectory objectives assign different importance to the same vision--language model (VLM) structures.
In the action expert, MLP channels exhibit more stable importance rankings across denoising steps than query heads, and a small fraction of channels accounts for most of the measured importance.
Based on these findings, we propose \ours{}, which combines importance ranks from both objectives for VLM pruning and aggregates stepwise importance to select a fixed subset of expert MLP channels.
At 24.0\% total parameter reduction, VLM-only pruning with \ours{} performs comparably to or slightly better than the strongest baselines in driving, while retaining 94.0\% of the dense model's LingoQA score, far above all other methods. At a higher reduction of 30.0\%, \ours{} degrades much less than every other method, as it is the only one whose reasoning does not collapse.
Removing 93.75\% of expert MLP channels increases total parameter reduction to 39.4\%, with nearly unchanged open-loop errors relative to VLM-only pruning and competitive closed-loop performance.
The resulting model achieves a $1.37\times$ end-to-end inference speedup and 36.6\% lower peak memory than the dense model, without post-pruning fine-tuning.

\end{abstract}

\section{Introduction}
\label{sec:intro}

Reasoning vision--language--action~(VLA) models for autonomous driving~\citep{nvidia2025alpamayo, zhou2025autovla, fu2025orion, yuan2026autodriver} extend action prediction with explicit chain-of-thought~(CoT) reasoning~\citep{wei2022chain} to understand complex driving situations and explain driving decisions.
Alpamayo~\citep{nvidia2025alpamayo}, for example, conditions a flow-based~\citep{lipman2023flow} action expert on context from its backbone, which includes Chain-of-Causation (CoC) reasoning generated by a vision--language model (VLM).
Incorporating CoC in action denoising process improves trajectory prediction and closed-loop driving over a trajectory-only baseline, particularly in challenging scenarios~\citep{nvidia2025alpamayo}.
However, this procedure involves substantial computation: a large VLM generates CoC tokens autoregressively, and an action expert runs repeatedly to predict trajectories through iterative denoising.
As autonomous driving requires continuous action updates under rapidly changing scene conditions, this computational bottleneck presents a critical challenge for real-time deployment~\citep{li2026flashdrive}.

\begin{figure*}[t]
    \centering
    \includegraphics[width=0.70\textwidth]{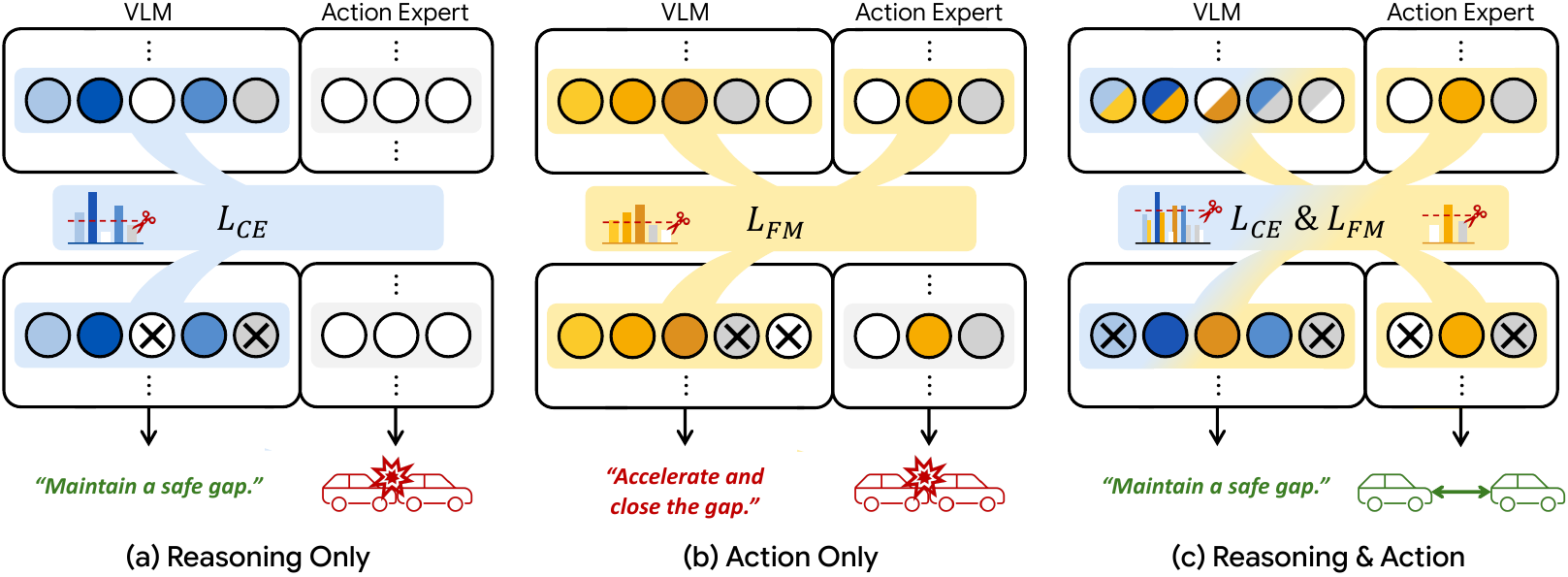}
    \caption{
    \textbf{Overview of Mixture-of-Objective Pruning (\ours).}
    (a) Pruning the VLM with reasoning-driven (CE) importance alone preserves the CoC but degrades the predicted trajectory.
    (b) Pruning with action-driven (FM) importance alone corrupts the reasoning that conditions the action expert.
    (c) \ours{} combines both objectives to select VLM units and uses FM importance to prune the action expert, preserving both reasoning and driving.
    }
    \label{fig:overview}
    \vspace{-0.5cm}
\end{figure*}

Structured model pruning offers a promising direction to reduce this cost by removing unnecessary parameters based on objective-driven importance scores~\citep{ma2023llm, li2025trthepruner, muralidharan2024compact}.
However, applying structured pruning to reasoning VLA is particularly challenging.
Unlike standard VLM, reasoning VLA leverages a shared VLM backbone for two different roles~(\textit{i.e.,} scene-level CoC reasoning and action expert conditioning), while the action expert operates under different mechanism~(\textit{i.e}., iterative denoising steps).
Deciding what to prune thus requires assessing structural importance across both training objectives and architectural components.

To derive a tailored pruning strategy for reasoning VLA, we examine Alpamayo~\citep{nvidia2025alpamayo}, a representative VLA framework for autonomous driving that predicts trajectories along with CoC reasoning.
Specifically, we assess the importance of structural units for both reasoning and action predictions using Taylor importance~\citep{molchanov2017pruningconvolutionalneuralnetworks,Molchanov_2019_CVPR, lecun1989optimal, 298572}, which estimates the loss change induced by removing each unit.
We further analyze the distribution of importance scores within the action expert across iterative denoising steps, accounting for its distinct operating mechanism from the autoregressive VLM backbone.

Our analysis of the VLM backbone reveals that the parameters important for reasoning and those important for action generation are only partially aligned.
Thus, pruning based solely on a single objective leads to either CoC collapse or a sharp decline in action prediction, as described in Figure~\ref{fig:overview}.
These findings suggest that preserving both scene understanding and action generation requires accounting for both objectives during pruning, unlike existing single-objective pruning methods.

For the action expert, parameter importance may vary across denoising steps.
We find that the importance of attention heads shifts with the step, whereas MLP channel importance remains nearly constant.
Moreover, while the importance score for Q-heads is broadly distributed, Figure~\ref{fig:expert_cumulative} discloses that the top 10\% of MLP channels account for approximately 98.2\% of the measured importance.

Motivated by these findings, we propose Mixture-of-Objective Pruning (MoOP), which jointly considers the dual objectives and the distinct architectural characteristics of both VLA modules.
For the VLM backbone, where the two objectives are only partially aligned, \ours{} scores each unit by the maximum of its importance ranks under the two objectives and removes units critical to neither CoC reasoning nor trajectory generation.
In the action expert, \ours{} instead prunes dense MLP channels that contribute little to prediction across denoising steps.
At a matched 24.0\% total parameter reduction, \ours{} matches or outperforms the compared methods on open- and closed-loop driving, and achieves the highest driving QA score by a large margin. At a higher reduction of 30.0\%, the gap widens: \ours{} is significantly better than all other methods in open-loop driving and the best in closed-loop driving.
With additional MLP-channel pruning in the action expert, the model reaches 39.4\% total parameter reduction, a $1.37\times$ end-to-end inference speedup and 36.6\% lower peak memory without post-pruning fine-tuning, while retaining competitive driving performance.

Our main contributions are as follows:
\begin{itemize}
    \item We demonstrate that VLM pruning must simultaneously account for both reasoning and action generation to preserve scene understanding and driving capability in reasoning VLA models.
    \item We discover that a small subset of MLP channels in the action expert is sufficient to maintain robust action prediction quality.
    \item We propose \ours{}, the only compared pruning method that preserves reasoning, which keeps it robust at higher pruning ratios where existing methods degrade sharply.
\end{itemize}

\section{Related Work}
\label{sec:related}

\textbf{Reasoning Vision--Language--Action Models for Autonomous Driving.}
Driving VLAs connect scene understanding and language reasoning with action prediction. DriveVLM~\citep{DBLP:conf/corl/TianGLLWZZJLZ24} integrates scene analysis with hierarchical planning, while ORION~\citep{fu2025orion} conditions a generative planner on an LLM-generated planning token. AutoVLA~\citep{zhou2025autovla} unifies reasoning and discrete actions within an autoregressive model. Alpamayo~\citep{nvidia2025alpamayo} generates Chain-of-Causation (CoC) reasoning and conditions a separate action expert on the VLM context to predict continuous trajectories.
We study how to prune this architecture while preserving its contributions to reasoning and trajectory prediction.

\textbf{Pruning Language and Vision-Language Models.}
Language-model pruning methods estimate importance through gradients or weight–activation statistics, as in LLM-Pruner~\citep{ma2023llm} and Wanda~\citep{sun2024simple}, and optimize layerwise sparsity allocation, as in Týr-the-Pruner~\citep{li2025trthepruner}. For VLMs, SlimVLM~\citep{wen2026slimvlm} accounts for visual-token redundancy and module sensitivity. Reasoning-aware approaches further adapt compression to the generation process: RAC~\citep{lucas2026reasoning} reconstructs activations from model-generated CoT~\citep{wei2022chain} traces, while MuCRASP~\citep{dutta2026mucrasp} considers reasoning-critical transitions and cross-modal alignment.
Our work examines a shared VLM that supports both textual reasoning and continuous action generation, combining importance measured under the two objectives.

\textbf{VLA Compression.}
VLA compression has explored structural pruning, quantization, and inference acceleration. For robotic manipulation, EfficientVLA~\citep{yang2025efficientvla} combines layer pruning, token selection, and action-head caching; CLP~\citep{nguyen2026finetuning} removes redundant VLM and action-head layers before fine-tuning; and QVLA~\citep{xu2026qvla} allocates channelwise precision using action sensitivity. In autonomous driving, Depth-Wise Probing and Pruning~\citep{babu2026depth} removes ORION decoder layers based on changes in the planning token, while FlashDrive~\citep{li2026flashdrive} accelerates Alpamayo through caching and speculative decoding.
We combine CoC- and trajectory-based VLM pruning with action-expert MLP pruning guided by importance stability and concentration across denoising steps, without post-pruning fine-tuning.

\section{Preliminaries}
\label{sec:preliminaries}

\subsection{Alpamayo - Reasoning VLA for Autonomous Driving}
\label{sec:reasoning_vla}

\begin{wrapfigure}{r}{0.42\textwidth}
    \centering
    \vspace{-0.45cm}
    \includegraphics[width=\linewidth]{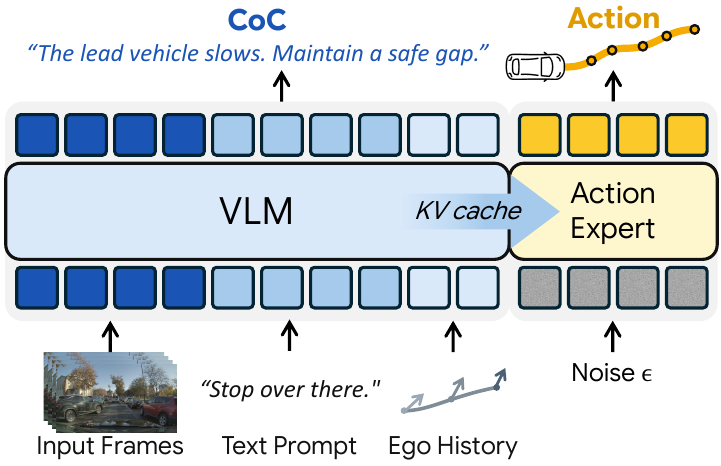}
    \caption{\textbf{Alpamayo architecture.} The VLM encodes frames, prompt and ego history, generates the CoC, and its KV cache conditions the action expert, which denoises noise $\epsilon$ into a trajectory.}
    \label{fig:architecture}
    \vspace{-0.4cm}
\end{wrapfigure}

Alpamayo~\citep{nvidia2025alpamayo} combines a reasoning VLM backbone~\citep{nvidia2025cosmosreason1physicalcommonsense} with a flow-based~\citep{lipman2023flow} action expert (Figure~\ref{fig:architecture}). Given multimodal input $x=(\mathcal I,e,p)$ which consists of multi-camera images $\mathcal I$, ego history $e$, and a text prompt $p$, the VLM autoregressively generates a Chain--of--Causation (CoC) $y=(y_1,\ldots,y_T)$ that links a driving decision to its causal factors. The VLM thus processes one sequence of images, ego-history and prompt text tokens followed by CoC tokens.
In each layer $\ell$ of the action expert, action tokens attend jointly to their own keys and values and to the full key--value (KV) cache of VLM layer $\ell$.
Conditioned on $\mathbf h_\theta(x,y)=\{(K_\ell,V_\ell)\}_{\ell=0}^{L-1}$, the expert denoises Gaussian noise in $S$ steps into acceleration and curvature values, converted into a future ego trajectory.

\textbf{Training objectives.}
The VLM $p_\theta$ is trained with cross-entropy (CE) on a ground-truth CoC $y$, and the action expert $v_\phi$ with flow matching (FM)~\citep{lipman2023flow} on a normalized ground-truth action $a$, using $z_t=(1-t)\epsilon+ta$ with $\epsilon\sim\mathcal N(0,I)$, $t\sim\mathcal U(0,1)$ and target velocity $v_t^*=a-\epsilon$:
\begin{equation}
    \mathcal L_{\mathrm{CE}}
    = -\frac{1}{T}\sum_{j=1}^{T}\log p_\theta(y_j\mid x,y_{<j}),
    \qquad
    \mathcal L_{\mathrm{FM}}
    =\mathbb E_{t,\epsilon}
    \left\|v_\phi(z_t,t;\mathbf h_\theta(x,y))-v_t^*\right\|_F^2.
    \label{eq:loss}
\end{equation}

\subsection{Structured Pruning and Taylor Importance}
\label{sec:taylor_importance}

\textbf{Structured pruning.}
Structured pruning reduces model size by removing computational units with low estimated importance, together with their parameters~\citep{ma2023llm, li2025trthepruner, muralidharan2024compact}. We prune two kinds of units, attention query heads and intermediate MLP channels, in both the VLM and the action expert. Removing query head $u$ deletes its $d_h$ rows of $W^{\mathrm Q}_\ell$ and the matching columns of $W^{\mathrm O}_\ell$; the key and value projections are kept, since several query heads share one KV head under grouped-query attention~\citep{ainslie2023gqa}. Removing MLP channel $u$ deletes row $u$ of $W^{\mathrm{up}}_\ell$ and $W^{\mathrm{gate}}_\ell$ and column $u$ of $W^{\mathrm{down}}_\ell$. Units are ranked by importance within each layer and unit type, the lowest-ranked ones are removed according to the budget, and the remaining weights are left unchanged. Why we choose query heads and intermediate MLP channels as pruning units is explained in Appendix~\ref{app:pruning_units}.

\textbf{Taylor importance.}
First-order Taylor importance estimates how much a loss changes when a unit is removed~\citep{molchanov2017pruningconvolutionalneuralnetworks,Molchanov_2019_CVPR}. We attach a gate $\alpha_{\ell,u}=1$ to the output of each unit: the head's attention output before $W^{\mathrm O}_\ell$, or the channel's activation before $W^{\mathrm{down}}_\ell$. Removing the unit sets $\alpha_{\ell,u}=0$, which to first order changes the loss on clip $c$ by $-\partial\mathcal L_c/\partial\alpha_{\ell,u}$. The importance of the unit under a loss $\mathcal L$ is

\medskip\noindent
\begin{minipage}{0.5\linewidth}
\begin{equation}
    \frac{\partial\mathcal L_c}{\partial\alpha_{\ell,u}}
    =\sum_{p}\mathbf o_{\ell,u,p}^{\top}
      \frac{\partial\mathcal L_c}{\partial\mathbf o_{\ell,u,p}},
    \label{eq:taylor_change}
\end{equation}
\end{minipage}%
\begin{minipage}{0.5\linewidth}
\begin{equation}
    I_{\mathcal L}(\ell,u)
    =\frac{1}{N}\sum_{c=1}^{N}
      \left|\frac{\partial\mathcal L_c}{\partial\alpha_{\ell,u}}\right|,
    \label{eq:taylor_importance}
\end{equation}
\end{minipage}\par\medskip
where $\mathbf o_{\ell,u,p}$ is the unit's output at token position $p$, the sum runs over all positions the unit processes, and $N$ is the number of calibration clips.
We write $I_{\mathrm{CE}}$ and $I_{\mathrm{FM}}$ for the importance measured under $\mathcal L_{\mathrm{CE}}$ and $\mathcal L_{\mathrm{FM}}$ (Eq.~\ref{eq:loss}), respectively. A unit with larger importance is estimated to change the loss more when it is removed, so the units with the smallest importance are pruned first.

\section{Analysis}
\label{sec:analysis}

We analyze Alpamayo-1.5-10B~\citep{nvidia2025alpamayo} to examine how importance is distributed in the VLM backbone under the two objectives and in the action expert across denoising steps.
We use 100 calibration clips selected from the PhysicalAI-AV training dataset~\citep{nvidia2025alpamayo} for Taylor importance computation. 
Details on calibration dataset sampling are in Appendix~\ref{app:calibration_sampling}.
For VLM units, $I_{\mathrm{FM}}$ is obtained by backpropagating the FM loss of the action expert through the KV cache into the VLM.
Importance is computed separately for the query heads and MLP channels of each layer, and all comparisons in this section are made within a layer and unit type, matching how units are selected for pruning in Section~\ref{sec:structured_pruning}.

\subsection{Importance Score in VLM Backbone}
\label{sec:vlm_analysis}

\begin{finding}
\textbf{Finding 1.} CE and FM importance rankings diverge in the late VLM layers. Even for the same generated CoC tokens, the two objectives favor different units for reasoning generation and action conditioning.
\end{finding}

\begin{figure}[t]
    \centering
    \begin{subfigure}[t]{0.32\linewidth}
        \centering
        \includegraphics[width=\linewidth]{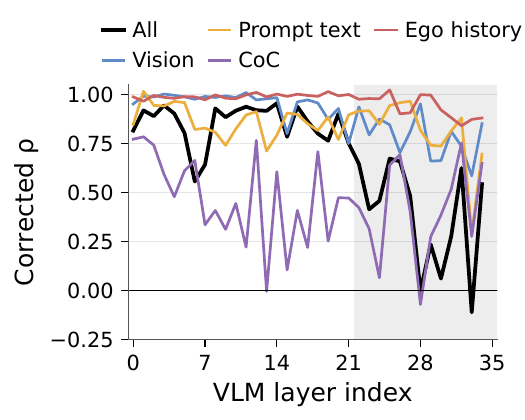}
        \caption{Rank correlation}
        \label{fig:vlm_rank_correlation}
    \end{subfigure}
    \hfill
    \begin{subfigure}[t]{0.32\linewidth}
        \centering
        \includegraphics[width=\linewidth]{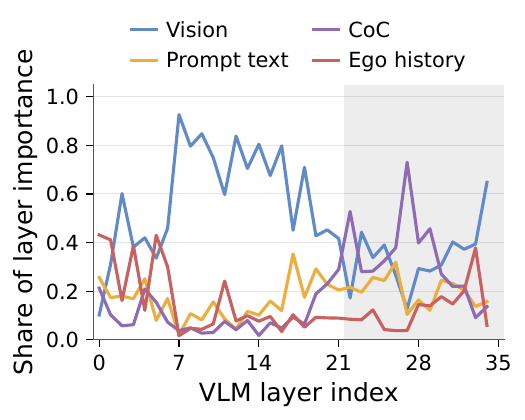}
        \caption{$I_{\mathrm{FM}}$ by token type}
        \label{fig:vlm_traj_share}
    \end{subfigure}
    \hfill
    \begin{subfigure}[t]{0.32\linewidth}
        \centering
        \includegraphics[width=\linewidth]{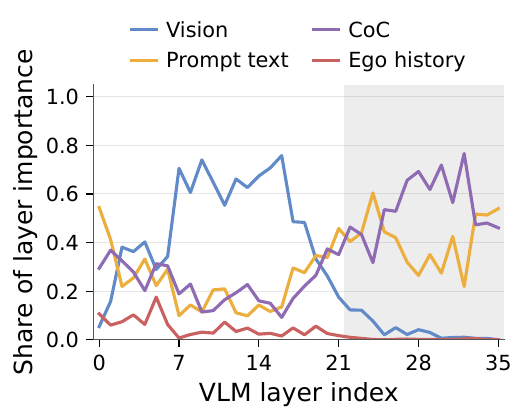}
        \caption{$I_{\mathrm{CE}}$ by token type}
        \label{fig:vlm_coc_share}
    \end{subfigure}
    \vspace{-0.1cm}
    \caption{\textbf{CE and FM importance differ in the late VLM layers.}
    (a) Rank correlation between the two objectives using all tokens (black) or one token type at a time (colored).
    (b, c) Layerwise importance shares for FM and CE, respectively, grouped into vision, prompt text, ego-history and CoC tokens.}
    \label{fig:vlm_importance}
    \vspace{-0.4cm}
\end{figure}

\begin{figure}[t]
    \centering
    \begin{subfigure}[t]{0.48\linewidth}
        \centering
        \includegraphics[height=6.5cm]{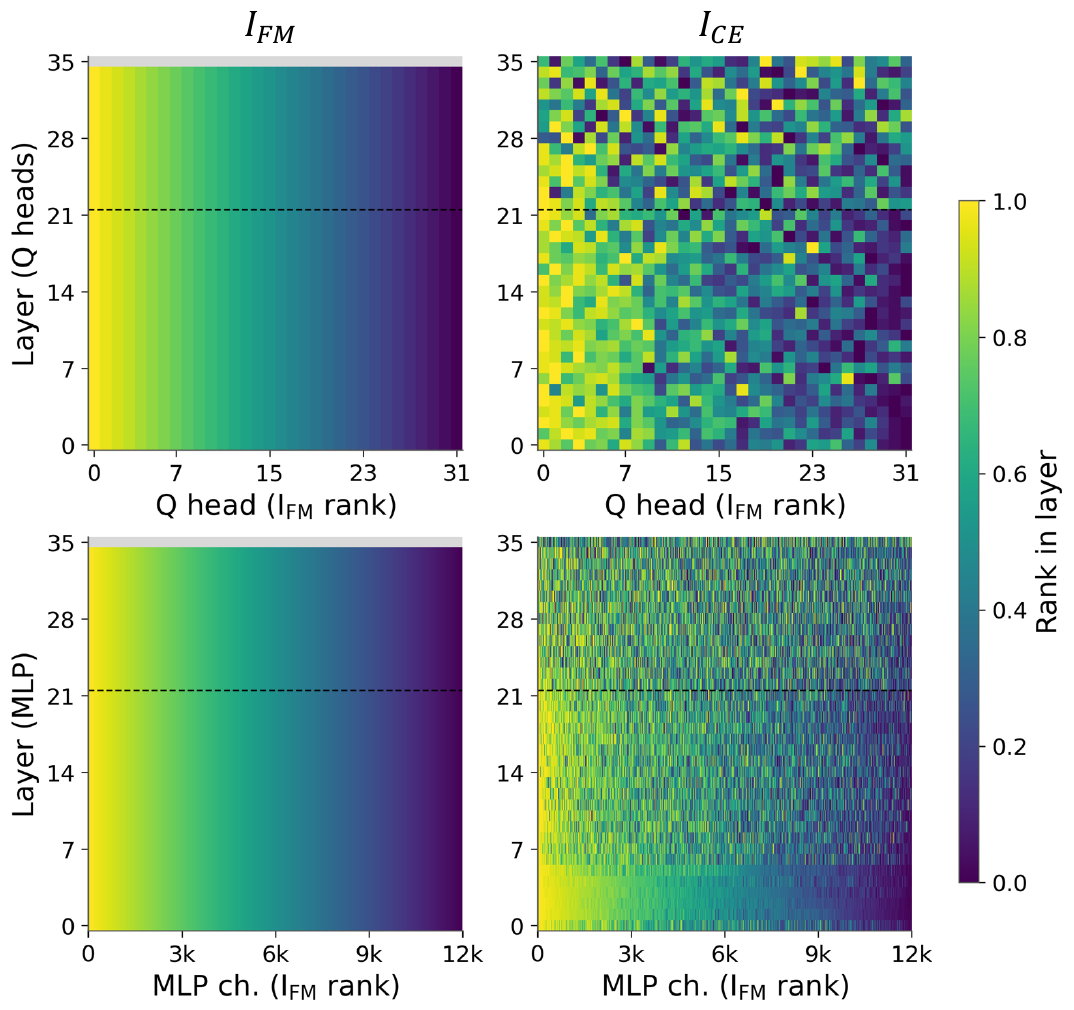}
        \caption{Rank within layer}
        \label{fig:rank_compare}
    \end{subfigure}
    \hfill
    \begin{subfigure}[t]{0.48\linewidth}
        \centering
        \includegraphics[height=6.5cm]{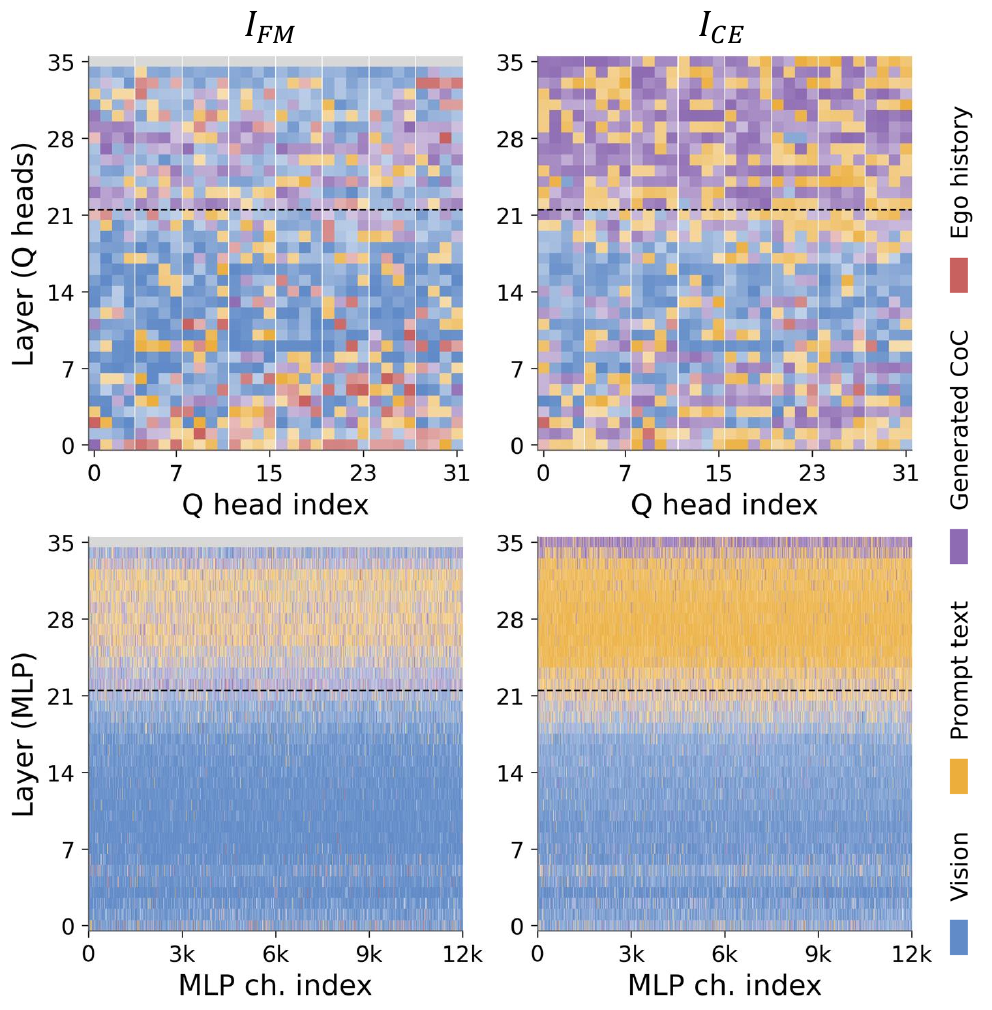}
        \caption{Dominant token type}
        \label{fig:vlm_head_token_map}
    \end{subfigure}
    \vspace{-0.1cm}
    \caption{\textbf{CE and FM importance differ for both query heads and MLP channels.}
    Each cell is a query head (top) or an MLP channel (bottom), with layers 0 to 35 from bottom to top.
    (a) Rank of each unit within its layer, ordered by $I_{\mathrm{FM}}$ rank; the closer the $I_{\mathrm{CE}}$ panel is to the $I_{\mathrm{FM}}$ ramp, the more the two rankings agree.
    (b) Token type contributing most to each unit's importance; paler colors mean the importance is spread over several types.
    }
    \label{fig:vlm_analysis}
    \vspace{-0.2cm}
\end{figure}

We first ask whether CE and FM importance identify the same VLM units for pruning. Within each layer, we compare the rankings induced by $I_{\mathrm{CE}}$ and $I_{\mathrm{FM}}$ using Spearman's $\rho$ (Figure~\ref{fig:vlm_rank_correlation}). 
The rankings largely agree through layer~21 but diverge in later layers, suggesting that pruning with either objective alone can remove units important to the other. Figure~\ref{fig:rank_compare} shows this comparison at the level of individual units.
The units of each layer are sorted by their $I_{\mathrm{FM}}$ rank, so each row of the $I_{\mathrm{FM}}$ panel changes smoothly from the most to the least important unit.
For both query heads and MLP channels, the $I_{\mathrm{CE}}$ panel departs from this order: units that $I_{\mathrm{FM}}$ ranks among the most important often receive low CE ranks, and vice versa.
To distinguish disagreement across token types from disagreement at the same tokens, we also compute importance separately for vision, prompt-text, generated-CoC and ego-history tokens by restricting the token-wise sum in the gate gradient (Eq.~\ref{eq:taylor_change}) to each token type before computing the Taylor importance score (Eq.~\ref{eq:taylor_importance}). The resulting rankings agree more closely for vision tokens, especially through layer~21, whereas agreement for generated CoC tokens remains low across much of the backbone. Thus, the disagreement is not solely due to the two objectives drawing importance from different token types.

We attribute this pattern to different token-wise importance distributions, as shown in Figure~\ref{fig:vlm_traj_share} and \ref{fig:vlm_coc_share}.
Both objectives place substantial importance on processing visual inputs in the early and middle layers, even though they may not rely on identical parameters, as indicated by their differing importance rank distributions.
However, in the late layers, we observe a notable difference in that CE importance becomes concentrated on text tokens, while FM importance remains distributed across vision, text and ego-history tokens. 
The query-head maps in Figure~\ref{fig:vlm_head_token_map} show this distinction at the level of individual heads: late-layer heads derive most of their CE importance from prompt-text or generated-CoC tokens, while their dominant sources of FM importance also include vision and ego-history tokens.

The persistent disagreement at generated CoC tokens reveals a further distinction between the objectives. The same unit outputs support next-token prediction under CE but provide conditioning information to the action expert through the KV cache under FM (Section~\ref{sec:reasoning_vla}). A unit needed to continue the reasoning sequence therefore need not be equally important for generating actions from that reasoning, and vice versa. CE and FM importance capture these different uses of the backbone, motivating a pruning criterion that accounts for both objectives.

\subsection{Importance Score in Action Expert}
\label{sec:expert_analysis}

\begin{finding}
\textbf{Finding 2.} Compared with query heads, MLP channels have more stable importance rankings across denoising steps and concentrate importance in fewer units.
\end{finding}

\begin{figure}[t]
    \centering
    \begin{subfigure}[t]{0.32\linewidth}
        \centering
        \includegraphics[width=\linewidth]{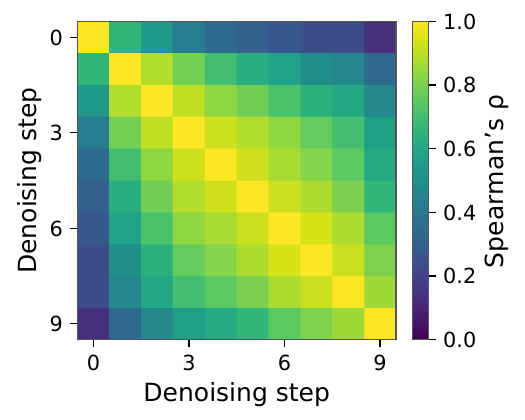}
        \caption{Q head rank similarity}
        \label{fig:expert_qhead_rank}
    \end{subfigure}
    \hfill
    \begin{subfigure}[t]{0.32\linewidth}
        \centering
        \includegraphics[width=\linewidth]{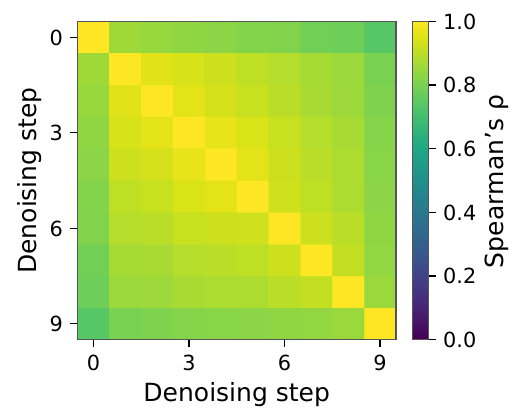}
        \caption{MLP channel rank similarity}
        \label{fig:expert_mlp_rank}
    \end{subfigure}
    \hfill
    \begin{subfigure}[t]{0.32\linewidth}
        \centering
        \includegraphics[width=\linewidth]{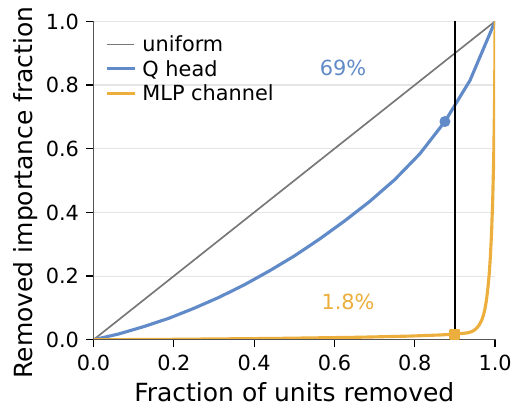}
        \caption{Cumulative importance}
        \label{fig:expert_cumulative}
    \end{subfigure}
    \caption{
        \textbf{Denoising stepwise rank stability and importance concentration in the action expert.}
        (a, b) Pairwise Spearman correlations of FM-based Taylor importance rankings across the ten denoising steps. Q head rankings become less similar between distant steps, whereas MLP channel rankings remain stable.
        (c) Fraction of the measured importance removed as units are removed in ascending order of importance. The lowest-scoring 90\% of MLP channels account for only 1.8\% of the measured importance, whereas Q head importance is distributed more broadly. The dashed diagonal represents a uniform importance distribution, and the vertical line marks 90\% removal.
    }
    \label{fig:expert_importance}
\end{figure}

The action expert reuses its weights across denoising steps, so removing a unit affects the entire generation process. 
We thus examine whether the same units remain important across steps. 
Let $I_{\mathrm{FM}}(\ell,u,s)$ denote the importance of unit $u$ in layer $\ell$, obtained by applying the Taylor criterion (Eq.~\ref{eq:taylor_importance}) to the FM loss evaluated at denoising step $s$. 
For each unit type, we compute the pairwise Spearman correlations between step-specific rankings and average them over layers (Figures~\ref{fig:expert_qhead_rank} and~\ref{fig:expert_mlp_rank}). 
To illustrate, we find that query-head rankings are consistent between nearby steps but diverge as the steps grow farther apart, while MLP-channel rankings are consistent across the steps.

We next examine how importance is distributed among units over the full denoising process. We summarize each unit's importance by averaging its step-specific scores:
\begin{equation}
    \bar I_{\mathrm{FM}}(\ell,u)
    =\frac{1}{S}\sum_{s=0}^{S-1}I_{\mathrm{FM}}(\ell,u,s),
    \label{eq:expert_mean_importance}
\end{equation}
where $S=10$ and the bar denotes averaging over denoising steps. Each step contributes equally with weight $1/S$, without additional timestep-dependent weighting. For each unit type, let $U_\ell$ be the number of units in layer $\ell$ and let $\pi_\ell(j)$ denote the index of the unit with the $j$-th smallest mean score. Thus, $\pi_\ell$ is a permutation of the unit indices satisfying
\[
    \bar I_{\mathrm{FM}}\bigl(\ell,\pi_\ell(1)\bigr)
    \leq\cdots\leq
    \bar I_{\mathrm{FM}}\bigl(\ell,\pi_\ell(U_\ell)\bigr).
\]
We define the cumulative importance fraction $C(r)$ as the fraction of total importance assigned to the lowest-scoring fraction $r$ of units, averaged over layers:
\begin{equation}
    C(r)=\frac{1}{L}\sum_{\ell=0}^{L-1}
      \frac{\sum_{j=1}^{\lfloor rU_\ell\rfloor}
                   \bar I_{\mathrm{FM}}\bigl(\ell,\pi_\ell(j)\bigr)}
           {\sum_{u=1}^{U_\ell}\bar I_{\mathrm{FM}}(\ell,u)},
    \label{eq:importance_concentration}
\end{equation}
where $L$ is the number of expert layers, $r\in[0,1]$ is the fraction of units considered for removal, and $\lfloor rU_\ell\rfloor$ is the corresponding integer count. A small $C(r)$ means that removing these units discards little of the measured importance. Figure~\ref{fig:expert_cumulative} shows that a large fraction of MLP channels accounts for little of the total importance, while query-head importance is distributed more broadly.

Together, these findings support targeting MLP channels for expert pruning: their stable rankings make a fixed selection of channels suitable across denoising steps, and their concentrated importance allows many low-scoring channels to be removed. Query heads offer less scope for this approach because their relative importance changes across steps and fewer heads have negligible scores.

\section{Method}
\label{sec:structured_pruning}

Our approach uses CE and FM importance jointly to select VLM units and step-averaged FM importance to select action expert MLP channels. 
In both components, selection is performed within each layer and unit type under a specified pruning budget. Applying the same pruning budget in every layer keeps comparisons focused on the scoring rule rather than on budget allocation. A non-uniform allocation guided by our analysis is explored in Appendix~\ref{app:analysis_expansion}.

\subsection{VLM Pruning with CE and FM Importance}
\label{sec:vlm_pruning}

Section~\ref{sec:vlm_analysis} shows that CE and FM importance can favor different VLM units. To account for both objectives without requiring their scores to have comparable scales, we combine their within-layer ranks. We apply the following rule separately to query heads and MLP channels, using the Taylor importance scores from Eq.~\ref{eq:taylor_importance}.

Let $U_\ell$ be the number of units of the selected type in layer $\ell$. Following the sorted-index notation in Section~\ref{sec:expert_analysis}, let $\pi_\ell^o(j)$ denote the index of the unit with the $j$-th smallest importance score under objective $o\in\{\mathrm{CE},\mathrm{FM}\}$. We define the corresponding ranks by
\begin{equation}
    \begin{gathered}
        I_o\bigl(\ell,\pi_\ell^o(1)\bigr)
        \leq\cdots\leq
        I_o\bigl(\ell,\pi_\ell^o(U_\ell)\bigr),\\
        \operatorname{rank}^{o}\bigl(\ell,\pi_\ell^o(j)\bigr)=j,
        \qquad j=1,\ldots,U_\ell.
    \end{gathered}
    \label{eq:importance_rank}
\end{equation}
Ties are broken by increasing unit index. The least important unit has rank $1$, and the most important has rank $U_\ell$.
We then define the combined rank score $R(\ell,u)$ as:
\begin{equation}
    R(\ell,u)
      =\max\!\left(\operatorname{rank}^{\mathrm{CE}}(\ell,u),
                   \operatorname{rank}^{\mathrm{FM}}(\ell,u)\right).
    \label{eq:vlm_combined_score}
\end{equation}
This maximum gives a unit high priority for retention if it is important to either objective; low maximum scores identify units that rank low under both. 
Since taking the maximum can give multiple units the same score, we note that $R$ is a pruning score rather than a unique ordering.
Alternative combination rules are compared in Appendix~\ref{app:ablations}.

For a target pruning ratio $r$, we sort units by increasing $R$ and remove the first $k_\ell(r)$ units:
\begin{equation}
    \begin{gathered}
        \sigma_\ell=\operatorname{argsort}_{u\in\{1,\ldots,U_\ell\}}R(\ell,u),
        \qquad k_\ell(r)=\left\lfloor rU_\ell+\tfrac{1}{2}\right\rfloor,\\
        \mathcal P_\ell(r)=\{\sigma_\ell(j):1\leq j\leq k_\ell(r)\}.
    \end{gathered}
    \label{eq:pruning_selection}
\end{equation}
Here, $\sigma_\ell(j)$ is the index of the unit at sorted position $j$, with score ties again broken by increasing unit index, and $\mathcal P_\ell(r)$ is the set of units to remove. All units outside $\mathcal P_\ell(r)$ are retained. 

\subsection{Action Expert MLP Pruning}
\label{sec:expert_pruning}

In the action expert, we prune intermediate MLP channels using FM importance. The CoC prediction loss depends on the VLM but not on the action expert parameters (Eq.~\ref{eq:loss}), so expert pruning requires only the FM criterion. Because the expert reuses its weights throughout denoising, we score each channel using the step-averaged importance $\bar I_{\mathrm{FM}}(\ell,u)$ in Eq.~\ref{eq:expert_mean_importance}. This score summarizes the estimated FM-loss sensitivity to removing a channel across the full denoising process. In practice, the choice of aggregation matters little. Using the score of any single denoising step instead yields nearly the same results (Appendix~\ref{app:analysis_expansion}), consistent with the stable rankings in Section~\ref{sec:expert_analysis}. 

The stable MLP importance rankings in Section~\ref{sec:expert_analysis} support retaining a fixed set of channels for all steps.
The concentration of MLP importance motivates evaluating high channel-removal ratios.
Within each expert layer, we first average the step-specific scores and then rank the channels once, applying Eq.~\ref{eq:importance_rank} to $\bar I_{\mathrm{FM}}$ with $o=\mathrm{FM}$.
We remove the lowest-ranked channels using the same selection rule as Eq.~\ref{eq:pruning_selection}, with $R(\ell,u)=\operatorname{rank}^{\mathrm{FM}}(\ell,u)$.
Expert attention is retained because query-head importance is more broadly distributed and its rankings vary across steps.
Pruning only the intermediate MLP channels preserves the residual-stream dimension and the layerwise KV-conditioning interface with the VLM, while reducing the MLP width at every denoising step.

\begin{table*}[t]
\centering
\caption{\textbf{Open-loop evaluation on PhysicalAI-AV~\citep{nvidia2025alpamayo}.} Rows with a 30\% pruning ratio remove 50\% of the units in each VLM layer ($r = 0.5$). 95\% confidence intervals are given in Appendices~\ref{app:reliability} and~\ref{app:higher_ratio}.}
\label{tab:open_loop}
\small
\setlength{\tabcolsep}{5pt}
\renewcommand{\arraystretch}{0.95}

\resizebox{\textwidth}{!}{
\begin{tabular}{l|c|c|cc|cc}
\toprule
 \multirow{2}{*}{Method}
& \multirow{2}{*}{\makecell{Expert\\Pruning}}
 &\multirow{2}{*}{\makecell{Pruning\\Ratio}}& \multicolumn{2}{c|}{Test500}
& \multicolumn{2}{c}{OOD-Val262} \\
 &  && minADE@6 (m) $\downarrow$ & minFDE@6 (m) $\downarrow$ & minADE@6 (m) $\downarrow$ & minFDE@6 (m) $\downarrow$ \\
\midrule
\midrule

 Baseline
& \xmark
 &0\%& 0.842 & 2.267 & 1.000 & 2.496 \\
\midrule

 Wanda
& \xmark
 &24\%& 2.975 & 8.269 & 2.790 & 6.884 \\
 LLM-Pruner
& \xmark
 &24\%& 1.451 & 3.889 & 1.662& 4.153 \\
 T\'yr-the-Pruner
& \xmark
 &24\%& \textbf{0.950} & 2.570& 1.177 & 2.990 \\
\midrule
 FM-only
& \xmark
 &24\%& 1.008 & 2.720 & 1.178 & 2.960\\
 CE-only
& \xmark
 &24\%& 1.584 & 4.500 & 1.546 & 4.010 \\

 \graycell \textbf{\ours{} (Ours)}& \graycell \xmark
 &\graycell 24\%& \graycell \textbf{0.950} & \graycell \underline{2.546}& \graycell \underline{1.119}& \graycell \underline{2.851}\\

 \graycell \textbf{\ours{} (Ours)}& \graycell \cmark
 &\graycell39.4\%& \graycell \underline{0.951}& \graycell \textbf{2.539}& \graycell \textbf{1.118}& \graycell \textbf{2.844}\\
\midrule
\midrule
Wanda & \xmark & 30\% & 2.901 & 7.869 & 2.761 & 6.917 \\
LLM-Pruner & \xmark & 30\% & 2.533 & 6.662 & 2.628 & 6.613 \\
T\'yr-the-Pruner & \xmark & 30\% & \underline{2.050} & \underline{5.339} & \underline{2.165} & \underline{5.202} \\
\midrule
FM-only & \xmark & 30\% & 2.822 & 8.278 & 2.436 & 6.753 \\
CE-only & \xmark & 30\% & 5.151 & 15.192 & 3.899 & 11.230 \\
\graycell \textbf{\ours{} (Ours)} & \graycell \xmark & \graycell 30\% & \graycell \textbf{1.237} & \graycell \textbf{3.338} & \graycell \textbf{1.416} & \graycell \textbf{3.616} \\
\bottomrule
\end{tabular}
}
\end{table*}

\begin{table}[t]
\centering
\caption{\textbf{Closed-loop evaluation in AlpaSim~\citep{alpasim_2025}.} Rows with a 30\% pruning ratio use $r = 0.5$. 95\% confidence intervals for all metrics and paired comparisons with MoOP are given in Appendices~\ref{app:reliability} and~\ref{app:higher_ratio}.}
\label{tab:closed_loop}
\small
\setlength{\tabcolsep}{5pt}
\renewcommand{\arraystretch}{0.95}
\resizebox{\textwidth}{!}{
\begin{tabular}{l|c|c|ccc|ccc}
\toprule
\multirow{2}{*}{Method}  &\multirow{2}{*}{\makecell{Expert\\Pruning}}& \multirow{2}{*}{\makecell{Pruning\\Ratio}}& \multicolumn{3}{c|}{Origin150} & \multicolumn{3}{c}{Hard100} \\
 &&& Scene Score $\uparrow$ & Offroad $\downarrow$& At-fault Coll. $\downarrow$ & Scene Score $\uparrow$ & Offroad $\downarrow$& At-fault Coll. $\downarrow$\\
\midrule
\midrule
Baseline & \xmark & 0\% & 0.750 & 0.070 & 0.040 & 0.510 & 0.265 & 0.030 \\
\midrule
Wanda &\xmark & 24\% & 0.621 & 0.093 & 0.147 & 0.406 & \textbf{0.140} & 0.060 \\
LLM-Pruner & \xmark & 24\% & \underline{0.811} & \textbf{0.037} & 0.057 & 0.575 & \underline{0.160} & 0.035 \\
Týr-the-Pruner & \xmark & 24\% & 0.786 & 0.087 & \underline{0.030} & 0.586 & 0.215 & \textbf{0.020} \\
\midrule
FM-only  &\xmark& 24\% & 0.783 & 0.087& 0.033 & 0.546 & 0.260 & \underline{0.025} \\
CE-only  &\xmark& 24\% & 0.660 & 0.090& 0.073 & 0.416 & 0.200 & 0.040 \\
\rowcolor{gray!15} \textbf{\ours{}(Ours)} & \xmark & 24\% & \textbf{0.828} & \underline{0.067} & \textbf{0.023}  & \underline{0.595} & 0.190 & 0.040 \\
\rowcolor{gray!15} \textbf{\ours{}(Ours)} & \cmark & 39.4\% & 0.793 & 0.083 & \underline{0.030} & \textbf{0.600} & 0.200 & \textbf{0.020} \\
\midrule
\midrule
Wanda & \xmark & 30\% & 0.350 & 0.280 & 0.180 & 0.266 & 0.290 & 0.025 \\
LLM-Pruner & \xmark & 30\% & 0.634 & \underline{0.073} & 0.143 & 0.454 & 0.210 & 0.055 \\
Týr-the-Pruner & \xmark & 30\% & \underline{0.742} & \underline{0.073} & \underline{0.087} & \underline{0.614} & \textbf{0.140} & \underline{0.020} \\
\midrule
FM-only & \xmark & 30\% & 0.551 & 0.103 & 0.097 & 0.289 & 0.240 & 0.070 \\
CE-only & \xmark & 30\% & 0.279 & 0.170 & 0.140 & 0.153 & 0.345 & 0.070 \\
\rowcolor{gray!15} \textbf{\ours{}(Ours)} & \xmark & 30\% & \textbf{0.866} & \textbf{0.047} & \textbf{0.037} & \textbf{0.651} & \underline{0.195} & \textbf{0.010} \\
\bottomrule
\end{tabular}
}
\end{table}

\section{Experiments}
\label{sec:experiments}

\subsection{Experimental Setup}
\label{sec:exp_setup}
\textbf{Implementation details.}
We prune Alpamayo-1.5-10B~\citep{nvidia2025alpamayo} using 100 calibration clips as illustrated in Section~\ref{sec:analysis}, without post-pruning fine-tuning.
Our main setting applies a VLM pruning ratio of $r=0.4$ in all 36 layers, removing 13 of 32 query heads and approximately $40\%$ of MLP channels, for $24.0\%$ total parameter reduction.
To test robustness under stronger compression, we also evaluate $r=0.5$, which removes half of the query heads and MLP channels in every VLM layer, for $30.0\%$ total reduction.
For the expert pruning process, we remove $93.75\%$ of MLP channels in each of its 36 layers, retaining 516 of $8{,}256$ channels and all attention heads, for $39.4\%$ total reduction on top of $r=0.4$.
At each ratio, all compared methods remove the same fraction of parameters.
Details of the baseline methods are provided in Appendix~\ref{app:compared_methods}.

\textbf{Evaluation protocol.}
Open-loop evaluation uses PhysicalAI-AV~\citep{nvidia2025alpamayo} Test500 and OOD-Val262, reporting the minimum average and final displacement errors over six 6.4\,s trajectories (minADE@6 and minFDE@6, in meters). Closed-loop evaluation uses AlpaSim~\citep{alpasim_2025} Origin150 and Hard100, reporting route-progress scene score with off-road and at-fault collision penalties, and the rates of these failures. Driving VQA uses LingoQA~\citep{marcu2024lingoqa}, reporting the percentage of answers accepted by Lingo-Judge. To measure how well CoC generation is retained, we also report the teacher-forced NLL of the CoC generated by the dense model on Test500 (CoC NLL). Higher scene score and Lingo-Judge score, and lower errors, failure rates, and CoC NLL, are better. Dataset construction and full protocols are in Appendix~\ref{app:evaluation_protocols}.
We compute 95\% confidence intervals with a cluster bootstrap and call a difference between two methods significant when its paired 95\% interval excludes zero (Appendix~\ref{app:reliability}).

\subsection{Main Results}
\paragraph{Open- and closed-loop driving.}
At $24.0\%$ parameter reduction, \ours{} outperforms CE-only and FM-only on all four open-loop metrics and both closed-loop scene scores (Tables~\ref{tab:open_loop} and~\ref{tab:closed_loop}).
On Hard100, its scene score exceeds the two variants by $0.179$ and $0.049$, respectively.
The improvement over CE-only is significant on both closed-loop sets, and that over FM-only is significant on Hard100 and over the combined 250 scenes (Appendix~\ref{app:reliability}).
Figure~\ref{fig:obj_comp_qual} in Appendix~\ref{app:closed_loop_qual} illustrates the corresponding driving behavior.
Joint importance thus benefits both trajectory prediction and driving under closed-loop feedback.

Against the external baselines, \ours{} is on par with the strongest methods at this budget.
It matches T\'yr-the-Pruner in Test500 minADE@6, lowers OOD-Val262 minADE@6 and minFDE@6 by $4.9\%$ and $4.6\%$, and achieves higher scene scores than both T\'yr-the-Pruner and LLM-Pruner on both closed-loop sets, although these closed-loop differences are not significant over the combined scenes.
Its advantage is clearer in long-tail scenarios: on OOD-train, a five-times larger set of OOD clips, \ours{} is significantly better than every other compressed method in open-loop prediction (Appendix~\ref{app:reliability}).

\begin{table}[t]
\centering
\caption{\textbf{Driving VQA evaluation on LingoQA~\citep{marcu2024lingoqa} and CoC NLL.} CoC NLL is computed on the CoC generated by the Baseline for Test500. Rows with a 30\% pruning ratio use $r = 0.5$. 95\% confidence intervals are given in Appendices~\ref{app:reliability} and~\ref{app:higher_ratio}.}
\label{tab:lingoqa_pruning}
\small
\setlength{\tabcolsep}{10pt}
\renewcommand{\arraystretch}{0.95}
\begin{tabular}{l|c|c|cc}
\toprule
Method & \makecell{Expert\\Pruning} & \makecell{Pruning\\Ratio} & LingoQA $\uparrow$ & CoC NLL $\downarrow$ \\
\midrule
\midrule
Baseline & \xmark & 0\% & 73.2 & 0.175 \\
\midrule
Wanda & \xmark & 24\% & 9.2 & 2.183 \\
LLM-Pruner & \xmark & 24\% & \underline{44.8} & 0.462 \\
T\'yr-the-Pruner & \xmark & 24\% & 34.2 & 0.516 \\
\midrule
FM-only & \xmark & 24\% & 37.0 & 0.586 \\
CE-only & \xmark & 24\% & 30.2 & \underline{0.313} \\
\rowcolor{gray!15}
\textbf{\ours{} (Ours)} & \xmark & 24\% & \textbf{68.8} & \textbf{0.252} \\
\midrule
\midrule
Wanda & \xmark & 30\% & 7.2 & 8.344 \\
LLM-Pruner & \xmark & 30\% & \underline{34.2} & 0.976 \\
T\'yr-the-Pruner & \xmark & 30\% & 6.2 & 2.254 \\
\midrule
FM-only & \xmark & 30\% & 1.6 & 1.852 \\
CE-only & \xmark & 30\% & 27.0 & \underline{0.540} \\
\rowcolor{gray!15}
\textbf{\ours{} (Ours)} & \xmark & 30\% & \textbf{47.4} & \textbf{0.376} \\
\bottomrule
\end{tabular}
\end{table}

\paragraph{Driving VQA evaluation.}
\ours{} exceeds CE-only and FM-only by $38.6$ and $31.8$ Lingo-Judge points, and the strongest external baseline, LLM-Pruner, by $24.0$ points (Table~\ref{tab:lingoqa_pruning}). It remains $4.4$ points below the dense model, a difference that is not significant, and achieves the lowest CoC NLL among the compressed models (Appendix~\ref{app:reliability}). The reasoning of the other methods, in contrast, collapses. A model that always answers ``yes'' already scores $37.6$ on LingoQA. FM-only and T\'yr-the-Pruner are statistically indistinguishable from this answer, CE-only and Wanda fall significantly below it, and LLM-Pruner exceeds it by only $7.2$ points. The baselines also fail on basic scene facts (Appendix~\ref{app:lingoqa_qual}). FM-only, T\'yr-the-Pruner and CE-only state that there is no traffic light where \ours{} reports its green state (Figure~\ref{fig:lq_signal}), and at a clear zebra crossing every compressed model except \ours{} answers incorrectly (Figure~\ref{fig:lq_zebra}b). CE-only is a notable case: it keeps the second-lowest CoC NLL but fails on LingoQA, because its importance is calibrated only on CoC text and therefore preserves CoC generation rather than general scene understanding. These results support considering both objectives to preserve driving-scene understanding.

\paragraph{Higher pruning ratio.}
The gap between \ours{} and the other methods widens under stronger compression.
At $r=0.5$ ($30.0\%$ total reduction), \ours{} achieves the highest scene score on both closed-loop sets, significantly above every other compressed method on Origin150 and above all but T\'yr-the-Pruner on Hard100 (Table~\ref{tab:closed_loop}).
Its open-loop errors are also significantly lower than those of every other method, with a Test500 minADE@6 of $1.237$\,m compared with $2.050$\,m for T\'yr-the-Pruner, the closest method (Table~\ref{tab:open_loop}).
In driving VQA, \ours{} keeps a Lingo-Judge score of $47.4$, significantly above all other methods and $13.2$ points above LLM-Pruner, the next best, which can no longer be distinguished from always answering ``yes''. \ours{} again has the lowest CoC NLL (Table~\ref{tab:lingoqa_pruning}).
Since the generated CoC conditions the action expert, this reasoning collapse is consistent with the sharper driving degradation of the baselines at this ratio.
Full results with confidence intervals are in Appendix~\ref{app:higher_ratio}.

\begin{table}[t]
\centering
\caption{\textbf{Action expert MLP pruning on \ours{}: open-loop errors, inference latency and memory usage.}}
\label{tab:mlp_removal_openloop}
\small
\setlength{\tabcolsep}{4pt}
\renewcommand{\arraystretch}{1.15}
\resizebox{\textwidth}{!}{%
\begin{tabular}{l c cc cc cccc cc}
\toprule
\multirow{2}{*}{Model} & \multirow{2}{*}{\makecell{Expert MLP\\Removed}} & \multicolumn{2}{c}{Test500 (m) $\downarrow$} & \multicolumn{2}{c}{OOD-Val262 (m) $\downarrow$} & \multicolumn{4}{c}{Latency} & \multicolumn{2}{c}{Memory} \\
\cmidrule(lr){3-4} \cmidrule(lr){5-6} \cmidrule(lr){7-10} \cmidrule(lr){11-12}
 & & minADE@6 & minFDE@6 & minADE@6 & minFDE@6 & \makecell{Prefill\\(ms)} & \makecell{Decode\\(ms/token)} & \makecell{Denoise\\(ms)} & \makecell{End-to-end\\(ms)} & \makecell{Params\\(B)} & \makecell{Peak\\(GiB)} \\
\midrule
Baseline & -- & 0.842 & 2.267 & 1.000 & 2.496 & 691.3 & 27.39 & 212.3 & 1341.8 (1.000$\times$) & 11.08 & 22.487 (1.000$\times$) \\
\midrule
\multirow{4}{*}{\ours} & 0\% & 0.950 & 2.546 & 1.119 & 2.851 & 515.6 & 19.53 & 204.3 & 1032.4 (1.300$\times$) & 8.42 & 17.448 (1.289$\times$) \\
 & 75\% & 0.950 & 2.543 & 1.118 & 2.848 & 514.4 & 19.52 & 160.3 & 987.0 (1.359$\times$) & 7.05 & 14.899 (1.509$\times$) \\
 & 87.5\% & 0.950 & 2.542 & 1.118 & 2.846 & 513.4 & 19.52 & 153.8 & 979.6 (1.370$\times$) & 6.82 & 14.473 (1.554$\times$) \\
 & 93.75\% & 0.951 & 2.539 & 1.118 & 2.844 & 514.1 & 19.54 & 150.9 & 977.8 (1.372$\times$) & 6.71 & 14.261 (1.577$\times$) \\
\bottomrule
\end{tabular}%
}
\end{table}

\paragraph{Action expert pruning.}
Additional expert pruning changes open-loop errors by at most $0.007$\,m across tested ratios (Table~\ref{tab:mlp_removal_openloop}). At $93.75\%$ MLP removal, the scene score changes by $-0.035$ on Origin150 and $+0.005$ on Hard100 (Table~\ref{tab:closed_loop}). This adds $15.4$ percentage points of parameter reduction while largely preserving driving performance.
This robustness depends on which channels are retained. Randomly selecting the same number of channels collapses at every ratio ($3.466$\,m Test500 minADE@6 at $93.75\%$ removal), so it is not explained by over-parameterization alone. Weight-magnitude selection performs comparably to our score on Test500 but is significantly worse on OOD-Val262 at the two highest ratios (Appendix~\ref{app:analysis_expansion}).

\subsection{Cost Analysis}

Combined pruning yields a $1.37\times$ end-to-end speedup and $36.6\%$ lower peak memory than the dense model (Table~\ref{tab:mlp_removal_openloop}).
The two components contribute complementary savings.
VLM pruning provides most of the end-to-end speedup ($1.30\times$), as it shortens prefill and autoregressive decoding, which dominate latency.
Expert pruning accelerates the denoising stage itself by $1.35\times$ ($204.3$ to $150.9$\,ms) and is essential for memory: it removes a further $1.71$B parameters and $3.19$\,GiB of peak memory, raising the peak-memory reduction from $22.4\%$ to $36.6\%$.
This saving relies on the stable MLP rankings of Section~\ref{sec:expert_analysis}. If the rankings changed across denoising steps, each step would need its own set of channels and the expert would have to keep every channel selected at any step, whereas here a single pruned expert serves all steps.

\section{Conclusion}
\label{sec:conclusion}

We presented \ours{}, a structured pruning method that accounts
for the roles of the VLM and action expert in reasoning VLAs.
Controlled comparisons show that using both reasoning and trajectory
importance better preserves driving QA, trajectory prediction,
and closed-loop scene scores than using either objective alone.
Among the compared pruning methods, \ours{} is the only one that preserves
driving-scene reasoning, and its advantage in driving grows at a higher pruning ratio.
For the action expert, we exploit the stability and concentration
of MLP importance across denoising steps to identify channels for removal.
Together, the two components compress Alpamayo-1.5-10B without any
post-pruning fine-tuning: VLM pruning provides most of the inference speedup,
and expert pruning further reduces the memory footprint while leaving
driving performance largely unchanged.
The two findings that motivate \ours{} also largely hold on Alpamayo 2 Edge
(Appendix~\ref{app:generalization}), suggesting that they are not specific to a single model.

\bibliography{main}
\bibliographystyle{iclr2027_conference}

\clearpage

\appendix

\newcommand{\SuppPanel}[5]{%
  \begin{minipage}[t]{#1}\vspace{0pt}\centering
    \includegraphics[
      page=#3,
      trim=#4,
      clip,
      width=\linewidth,
      height=#5\textheight,
      keepaspectratio
    ]{#2}%
  \end{minipage}
}
\newcommand{\SuppTag}[1]{%
  \makebox[\linewidth][l]{\small\textbf{(#1)}}\par\nointerlineskip\vspace{2pt}}
\newcommand{\SuppClosedCase}[4]{%
  \begin{figure}[p]
    \centering
    \includegraphics[page=#1,trim=#2,clip,width=\linewidth,
      height=0.78\textheight,keepaspectratio]{\SuppClosedPDF}
    \caption{#3}
    \label{#4}
  \end{figure}}

\newcommand{\SuppClosedCaseNp}[4]{%
  \begin{figure}[!h]
    \centering
    \includegraphics[page=#1,trim=#2,clip,width=\linewidth,
      height=0.6\textheight,keepaspectratio]{\SuppClosedPDF}
    \caption{#3}
    \label{#4}
  \end{figure}}

\providecommand{\SuppLingoPDF}{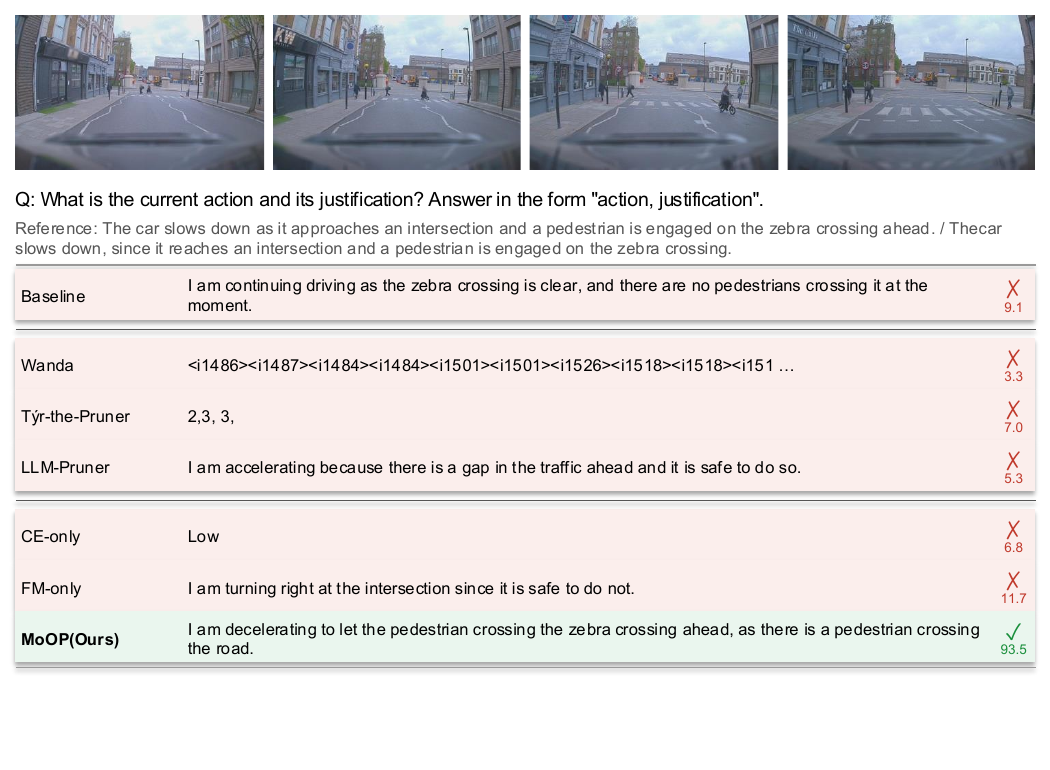}
\providecommand{\SuppOpenPDF}{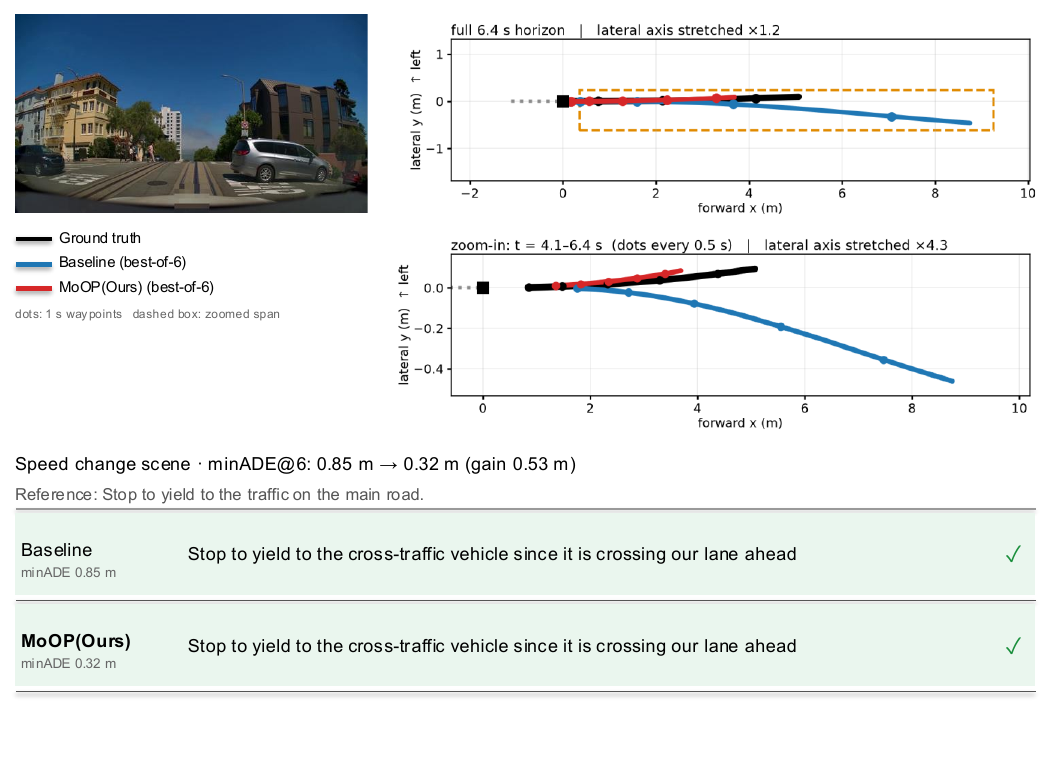}
\providecommand{\SuppClosedPDF}{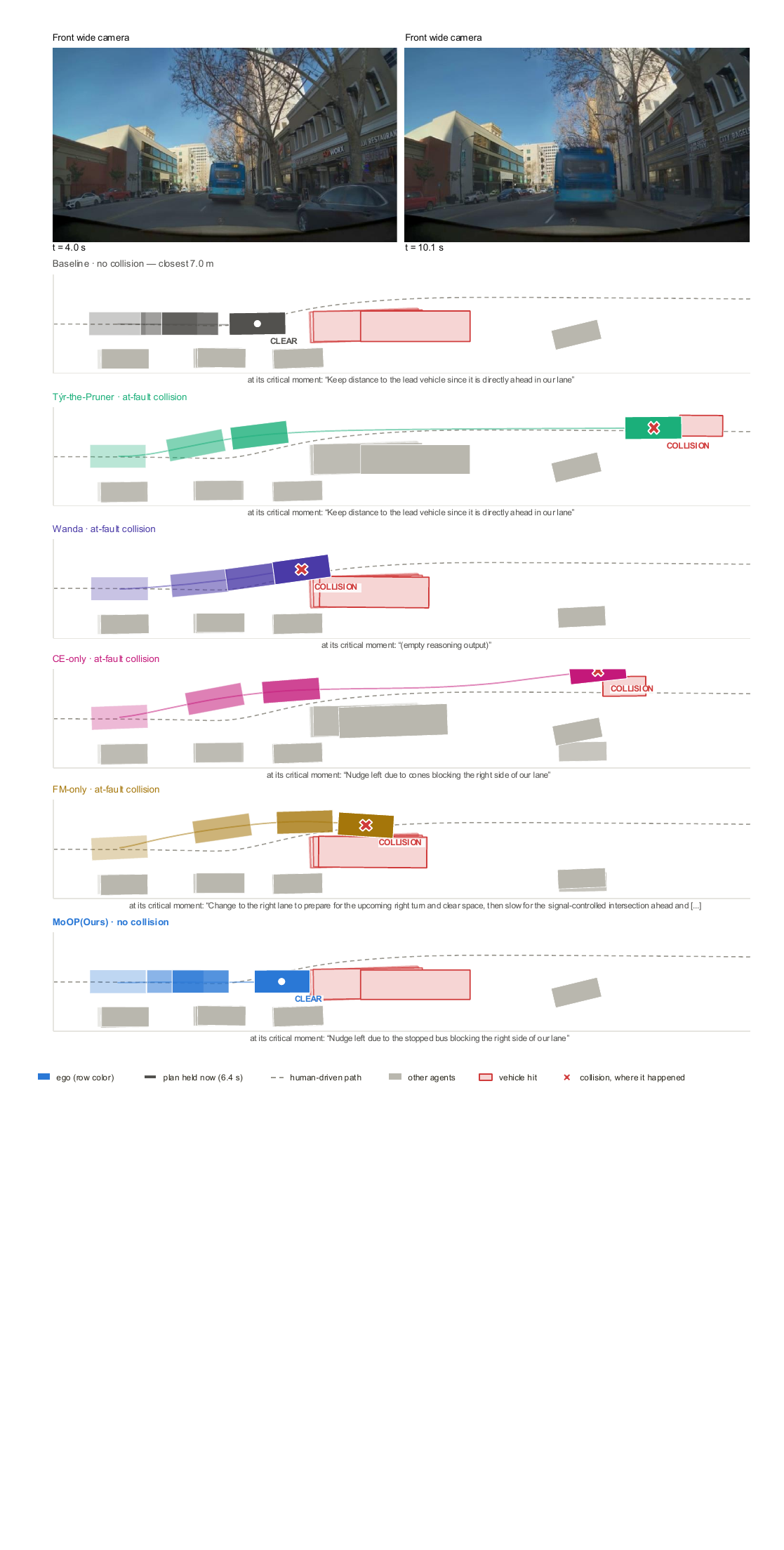}
\providecommand{\SuppRoutePDF}{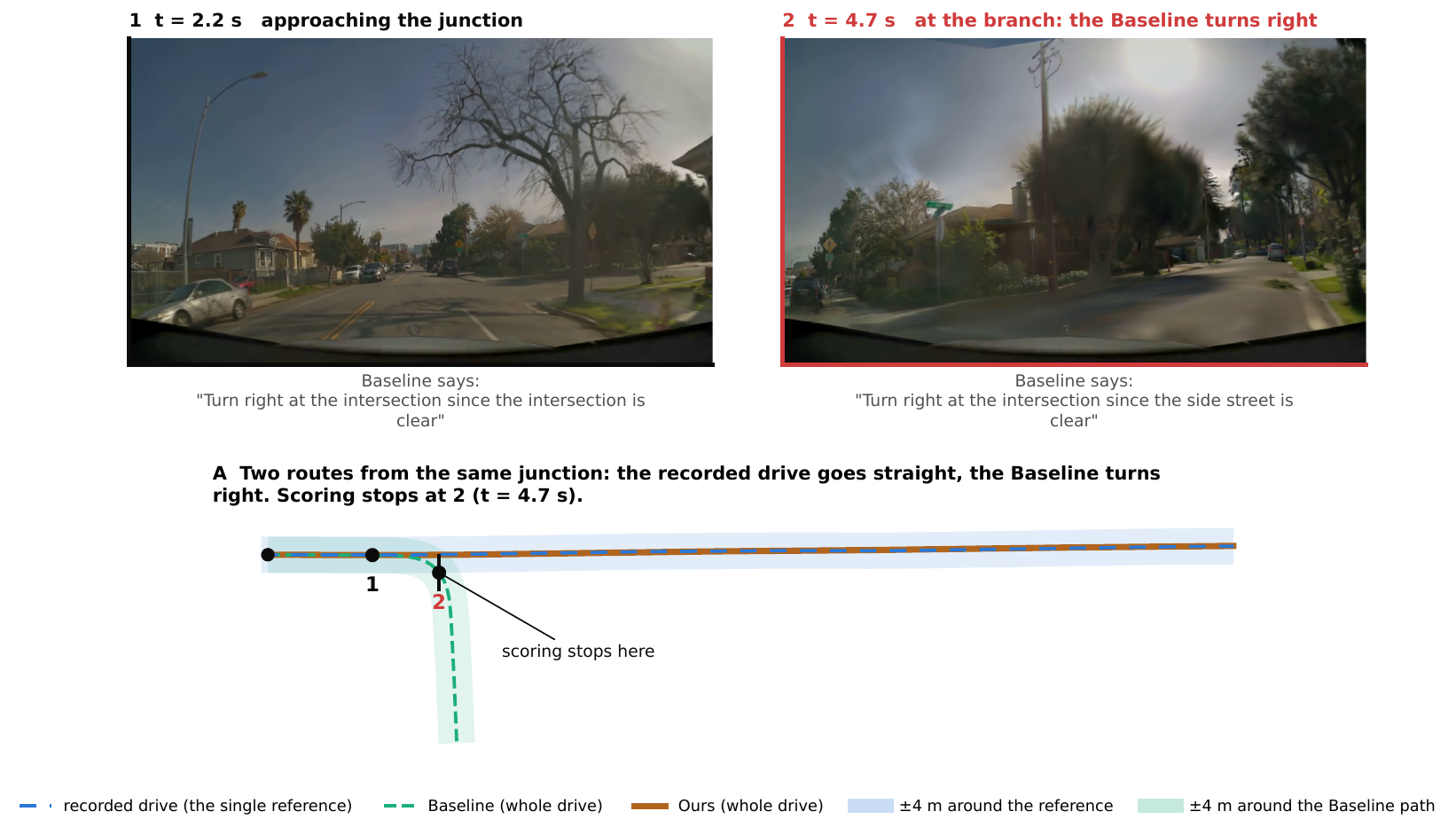}
\newcommand{\SuppRouteCase}[3]{%
  \begin{figure}[!h]
    \centering
    \includegraphics[page=#1,width=\linewidth,
      height=0.45\textheight,keepaspectratio]{\SuppRoutePDF}
    \caption{#2}
    \label{#3}
  \end{figure}}

\providecommand{\tbd}{\textcolor{red}{[x]}}
\providecommand{\cival}[2]{$#1$\,{\scriptsize$[#2]$}}
\providecommand{\valtbd}[1]{$#1$\,{\scriptsize\tbd}}
\providecommand{\civs}[2]{\begin{tabular}[c]{@{}c@{}}$#1$\\[-1pt]{\scriptsize$[#2]$}\end{tabular}}
\providecommand{\civssd}[3]{\begin{tabular}[c]{@{}c@{}}$#1$\,{\scriptsize$(#2)$}\\[-1pt]{\scriptsize$[#3]$}\end{tabular}}
\providecommand{\twoline}[2]{\begin{tabular}[c]{@{}l@{}}#1\\[-1pt]{\scriptsize #2}\end{tabular}}

\begin{center}
    {\large\bfseries Appendix\par}
\end{center}

\section*{Overview}
\label{sec:appendix_overview}

This appendix describes calibration data selection, compared methods, and the sampling and evaluation protocols for each benchmark.
It then reports confidence intervals and results at a higher pruning ratio, followed by additional analyses, ablations and a generalization study.
Qualitative results complement the quantitative evaluation in the main paper.
The content is organized as follows:

\begin{itemize}
    \item \textbf{Sec.~\ref{app:data_sampling}}: Experimental details.
    \begin{itemize}
        \item \textbf{Sec.~\ref{app:calibration_sampling}}: Construction of the \textit{Calib100} calibration set from the training split by greedy matching of collection-metadata distributions.
        \item \textbf{Sec.~\ref{app:compared_methods}}: External pruning baselines, single-objective variants, and expert-pruning settings.
        \item \textbf{Sec.~\ref{app:evaluation_protocols}}: Evaluation protocols and data sampling for open-loop trajectory prediction, closed-loop driving, and LingoQA.
        \item \textbf{Sec.~\ref{app:pruning_units}}: Rationale for pruning query heads and intermediate MLP channels.
    \end{itemize}
    \item \textbf{Sec.~\ref{app:quantitative_results}}: Quantitative results.
    \begin{itemize}
        \item \textbf{Sec.~\ref{app:reliability}}: Confidence intervals for the quantitative results in Sec.~\ref{sec:experiments}.
        \item \textbf{Sec.~\ref{app:higher_ratio}}: Robustness at a higher pruning ratio.
    \end{itemize}
    \item \textbf{Sec.~\ref{app:discussion}}: Additional experiments and discussion.
    \begin{itemize}
        \item \textbf{Sec.~\ref{app:analysis_expansion}}: Extended analysis: non-uniform layer-wise allocation, step-wise expert scoring, and expert channel selection.
        \item \textbf{Sec.~\ref{app:ablations}}: Ablation of the combination rule.
        \item \textbf{Sec.~\ref{app:generalization}}: Generalization to Alpamayo 2 Edge.
    \end{itemize}
    \item \textbf{Sec.~\ref{app:qualitative_results}}: Qualitative results.
    \begin{itemize}
        \item \textbf{Sec.~\ref{app:lingoqa_qual}}: Qualitative comparison on LingoQA driving question answering, grouped by signal state, road users and hazards, and actions at zebra crossings.
        \item \textbf{Sec.~\ref{app:open_loop_qual}}: Qualitative comparison of open-loop trajectory predictions and the reasoning generated with them, between the Baseline and \ours{}.
        \item \textbf{Sec.~\ref{app:closed_loop_qual}}: Qualitative comparison of closed-loop rollouts in AlpaSim, covering scenes where other pruning methods cause at-fault collisions or leave the road.
        \item \textbf{Sec.~\ref{app:multi_answer}}: Qualitative discussion of multi-answer cases in closed-loop evaluation, where a scene with more than one valid route is scored against a single recorded path.
    \end{itemize}
\end{itemize}

\section{Experimental Details}
\label{app:data_sampling}

\subsection{Calibration Set Selection}
\label{app:calibration_sampling}

We construct Calib100 from the official training split of the PhysicalAI-AV dataset~\citep{nvidia2025alpamayo}. Of the 153,625 training clips, we exclude
1,450 clips listed in the dataset's OOD reasoning annotations, leaving
152,175 candidates. We select 100 clips using seeded greedy matching
of collection-metadata distributions. The target distributions are
computed over the full training split before these exclusions.

The procedure matches the marginal distributions of six categorical
attributes: country, platform class, time of day, season, month, and
radar configuration, with respective weights of
$4.0$, $2.0$, $2.0$, $1.5$, $1.0$, and $1.0$.
For a selected set $\mathcal S$ and a reference split $\mathcal R$,
we measure their discrepancy as
\begin{equation}
    J_{\mathcal R}(\mathcal S)
    =
    \frac{1}{\sum_{a\in\mathcal A} w_a}
    \sum_{a\in\mathcal A} w_a
    \sum_{v\in\mathcal V_a}
    \left|
        \widehat p_{a,v}(\mathcal S)-p_{a,v}(\mathcal R)
    \right|,
    \label{eq:metadata_matching}
\end{equation}
where $\mathcal A$ is the attribute set, $\mathcal V_a$ is the set of
categories for attribute $a$, and $\widehat p_{a,v}(\mathcal S)$ and
$p_{a,v}(\mathcal R)$ are the corresponding category proportions.
Starting from an empty set, we repeatedly add the unselected candidate
that minimizes the resulting discrepancy. Ties are broken uniformly
at random using a NumPy~\citep{harris2020array} random generator initialized with seed 42.
Calib100 consists of the first 100 clips in this greedy sequence.

Time of day is classified as daytime for
$6 \leq \text{hour} < 18$ and nighttime otherwise.
Season is derived from calendar month: December--February for winter,
March--May for spring, June--August for summer, and September--November
for fall, using the same convention for all countries.
Missing values form a separate category for each attribute.
The selected set attains a weighted discrepancy of $0.0306$ under
Eq.~\ref{eq:metadata_matching}. Each clip uses a fixed window start
of $t_0=5.1$\,s.
Selection uses only collection metadata; driving actions and model
outputs are not used to choose clips.

\subsection{Compared Methods}
\label{app:compared_methods}

All compared models start from Alpamayo-1.5-10B. The dense model,
denoted Baseline, is evaluated without pruning. For the main VLM
comparison, all pruning methods remove $24.0\%$ of the full model's
parameters and leave the action expert intact. We evaluate every model
without post-pruning fine-tuning, so recovery training does not account
for differences between methods.

\paragraph{LLM-Pruner.}
LLM-Pruner~\citep{ma2023llm} identifies groups of dependent structures
and estimates their importance using gradient information. Removing
these groups together preserves the connectivity of the remaining
network. We apply this structural pruning baseline to the VLM and omit
the optional LoRA recovery stage used in the original method.

\paragraph{Wanda.}
Wanda~\citep{sun2024simple} scores a weight by the product of its absolute
magnitude and the norm of its input activations. The original method
removes low-scoring weights to produce unstructured sparsity. In our
comparison, we use Wanda's importance criterion to select structured
VLM units, enabling comparison with methods that physically remove
query heads and MLP channels.

\paragraph{T\'yr-the-Pruner.}
T\'yr-the-Pruner~\citep{li2025trthepruner} optimizes the distribution of
sparsity across layers under an overall parameter budget. It constructs
a supernet of locally pruned candidates at multiple sparsity levels and
uses a coarse-to-fine search to select their combination. This provides
a comparison with global sparsity allocation, whereas \ours{} uses a
uniform per-layer budget and combines importance from two objectives.
The total parameter reduction is matched in our VLM comparison.

\paragraph{Single-objective variants.}
CE-only and FM-only isolate the effect of the pruning objective. They
use $I_{\mathrm{CE}}$ and $I_{\mathrm{FM}}$, respectively, to rank units
within each layer and unit type, then remove the lowest-ranked units.
Both variants match \ours{} in the Taylor criterion, calibration set,
pruning targets, and per-layer removal counts. They therefore test
whether either objective alone suffices to preserve the VLM's reasoning
and action-conditioning roles.

\paragraph{\ours{} and additional expert pruning.}
\ours{} combines CE and FM ranks using the maximum-rank rule in
Section~\ref{sec:vlm_pruning}. We separately assess additional expert
MLP pruning on top of this VLM, using step-averaged FM importance as
described in Section~\ref{sec:expert_pruning}. The expert ablation removes
$75\%$, $87.5\%$, or $93.75\%$ of MLP channels while retaining expert
attention (Table~\ref{tab:mlp_removal_openloop}). The main configuration
uses $93.75\%$ expert MLP removal, reaching $39.4\%$ total parameter
reduction. Results with and without expert pruning are distinguished
in the main tables.

\subsection{Evaluation Protocols and Data Sampling}
\label{app:evaluation_protocols}

We evaluate trajectory prediction, closed-loop driving, and driving
question answering. Within each benchmark, all models receive the same
evaluation examples and use the same evaluation settings. Data selection
and metric computation are described together below for each benchmark.

\subsubsection{Open-Loop Trajectory Prediction}
\label{app:open_loop_sampling}

\paragraph{Test500.}
We select 500 clips from the official test split using the same metadata attributes, weights, greedy procedure, and seed as Calib100.
The reference distributions are computed over all 61,599 test clips. We exclude the 12 test clips of an earlier proof-of-concept calibration set, leaving 61,587 candidates. Test500 consists of the first 500 clips in the greedy sequence and attains a weighted discrepancy of $0.0055$ relative to the full test split.
As in calibration, each clip uses $t_0=5.1$\,s.
Test500 and Calib100 are clip-disjoint.

\paragraph{OOD-Val262.}
For out-of-distribution evaluation, we use all 262 clips in the
validation split provided by the PhysicalAI-AV~\citep{nvidia2025alpamayo} OOD sample, without
additional subsampling. OOD-Val262 is clip-disjoint from both
Calib100 and Test500.

\paragraph{Evaluation protocol and metrics.}
For each clip, the model generates six candidate ego trajectories over
6.4\,s from the recorded driving context. We compare these predictions
with the logged future trajectory without executing them in a simulator.
Let $\hat{\mathbf p}_{i,k,t}\in\mathbb R^2$ and $\mathbf p_{i,t}$ denote
the predicted and ground-truth positions in the same coordinate frame,
where $i$ indexes clips, $k$ indexes the six candidates, and $t$ indexes
$H$ future timestamps. For $N$ clips, we report
\begin{equation}
\begin{aligned}
    \operatorname{minADE@6}
    &=\frac{1}{N}\sum_{i=1}^{N}\min_{1\leq k\leq6}
      \frac{1}{H}\sum_{t=1}^{H}
      \left\|\hat{\mathbf p}_{i,k,t}-\mathbf p_{i,t}\right\|_2,\\
    \operatorname{minFDE@6}
    &=\frac{1}{N}\sum_{i=1}^{N}\min_{1\leq k\leq6}
      \left\|\hat{\mathbf p}_{i,k,H}-\mathbf p_{i,H}\right\|_2.
\end{aligned}
    \label{eq:open_loop_metrics}
\end{equation}
The minima are taken independently for the two metrics before averaging
over clips, so the selected candidates may differ. Both errors are
measured in meters, and lower values indicate closer agreement with the
recorded trajectory. We use $N=500$ for Test500 and $N=262$ for OOD-Val262.

\subsubsection{Closed-Loop Driving}
\label{app:closed_loop_sampling}

Both closed-loop evaluation sets are drawn from the 913 scenes in
AlpaSim's \texttt{public\_2601} release~\citep{alpasim_2025}. Scene artifacts are pinned
by UUID to this release to avoid substituting renders from later
releases.

\paragraph{Origin150.}
Origin150 contains 150 scenes randomly sampled from the parent suite.

\paragraph{Hard100.}
We construct a separate set of 100 scenes using a difficulty score
derived from ground-truth ego motion. From each scene's ego pose
track sampled at 10\,Hz, we compute the mean speed $\bar v_i$ and
the accumulated absolute yaw change $\sum_t |\Delta\psi_{i,t}|$.
The score for scene $i$ is
\begin{equation}
    h_i
    =
    z_{\mathcal P}(\bar v_i)
    +
    z_{\mathcal P}\!\left(\sum_t |\Delta\psi_{i,t}|\right),
    \label{eq:hard100_score}
\end{equation}
where $z_{\mathcal P}$ standardizes each feature using its mean
and standard deviation over the full parent suite $\mathcal P$
of 913 scenes. No model output enters this score.

We exclude all Origin150 scenes and any scene whose ground-truth
ego path is shorter than 5\,m. Among the remaining scenes, we retain
the 100 with the highest scores.
Hard100 is therefore disjoint from Origin150, and the two sets
together cover 250 scenes.
Hard100 uses two rollouts per scene, yielding 200 rollouts.
Its selection criterion emphasizes speed and turning demands
and does not explicitly stratify scenes by collision risk.

\paragraph{Evaluation protocol.}
In AlpaSim, the model repeatedly predicts actions from simulated observations, and the executed actions update the ego state in a scene reconstructed from recorded logs.
Each model is evaluated on the same scenes with two rollouts per scene, giving 300 rollouts for Origin150 and 200 for Hard100.
All rates in Table~\ref{tab:closed_loop} are computed over all rollouts of each set.
Metrics are averaged separately within each set; because every scene has two rollouts, scenes receive equal weight.

\paragraph{Metrics.}
We report scene score, off-road rate, and at-fault collision rate using
AlpaSim's evaluator~\citep{alpasim_2025}. Let $b_j$ and $c_j$ be binary
indicators of any off-road event and any at-fault collision in rollout
$j$, respectively. Multiple occurrences within a rollout count once.
AlpaSim's at-fault criterion includes front and lateral impacts and
excludes rear collisions. Let $q_j\in[0,1]$ be the normalized route-progress
score computed with AlpaSim's progress normalization and saturation rules.
For $M$ rollouts, the metrics are
\begin{equation}
\begin{aligned}
    \operatorname{SceneScore}
    &=\frac{1}{M}\sum_{j=1}^{M}q_j(1-b_j)(1-c_j),\\
    \operatorname{OffRoadRate}
    &=\frac{1}{M}\sum_{j=1}^{M}b_j,
    \qquad
    \operatorname{AtFaultCollisionRate}
    =\frac{1}{M}\sum_{j=1}^{M}c_j.
\end{aligned}
    \label{eq:closed_loop_metrics}
\end{equation}
An off-road event or at-fault collision sets the corresponding rollout's
scene score to zero. All metrics lie on a 0--1 scale; higher scene score
and lower failure rates are better. Route progress is measured against
the recorded path, which can penalize a different but valid route;
Appendix~\ref{app:multi_answer} illustrates such cases.

\subsubsection{Driving Question Answering on LingoQA}
\label{app:lingoqa_evaluation}

\paragraph{Evaluation data and inputs.}
We use the full official LingoQA evaluation split~\citep{marcu2024lingoqa}:
500 questions from 100 segments, with five questions per segment and two
human reference answers per question. No questions are subsampled or
filtered. Each segment provides five front-camera frames indexed 0--4.
We use the last four (1--4), closest to the question time, to match
Alpamayo-1.5's four-frame input per camera. All four are labelled as the
front camera, and the question is passed verbatim through the model's
native VQA template, without additional style instructions or ego-trajectory
inputs.

\paragraph{Evaluation protocol and metrics.}
Evaluation is zero-shot, with no training on LingoQA. Answers are decoded
greedily with a maximum of 128 tokens. We use the official Lingo-Judge
checkpoint, take the higher of the two reference-conditioned logits,
and count an answer as correct when this logit exceeds zero. The reported
score is the percentage of correct answers over all 500 questions.

\paragraph{CoC likelihood.}
Alongside LingoQA, we report the teacher-forced negative log-likelihood (NLL) of CoC text, averaged over tokens.
It is computed on the CoC generated by the Baseline for the Test500 clips, and measures how well a compressed model retains the CoC generation of the Baseline.
Appendix~\ref{app:reliability} additionally reports it on the ground-truth CoC of OOD-train.

\subsection{Choice of Pruning Unit Types}
\label{app:pruning_units}

Query heads and intermediate MLP channels define the granularity of pruning, that is, the type of unit that can be removed. We did not choose these unit types because they were estimated to be less important than other types, and we did not compare importance across unit types in advance. Importance is measured afterwards, only to decide which individual heads and channels to remove within each type. Likewise, other units, such as layers, hidden dimensions and KV heads, are kept not because they are more important, but because of the architecture. 

In Alpamayo~\citep{nvidia2025alpamayo}, each expert layer $l$ attends to the KV cache of VLM layer $l$, which is how the VLM conditions action generation. Removing layers or hidden dimensions would break this layer-to-layer conditioning, and removing KV heads would remove part of the KV cache itself, which is shared by several query heads under grouped-query attention~\citep{ainslie2023gqa}. Query heads and intermediate MLP channels, in contrast, can be removed without changing this interface. They are also the standard units in structured pruning of transformers~\citep{michel2019sixteen,voita2019analyzing,kwon2022fast,ma2023llm,muralidharan2024compact}. Removing them deletes entire rows and columns of the weight matrices, so the matrices simply become smaller. For example, removing an MLP channel deletes one row of the up and gate projections and one column of the down projection. Unlike removing individual weights, which leaves scattered zeros that standard hardware cannot skip efficiently, this yields real speedups without specialized sparse kernels.

\clearpage

\section{Quantitative Results}
\label{app:quantitative_results}

\subsection{Confidence Intervals for Main Results}
\label{app:reliability}

We add 95\% confidence intervals to the main tables in Section~\ref{sec:experiments}. The intervals are percentile intervals from a cluster bootstrap, which resamples groups of correlated samples together rather than individual samples. For closed-loop evaluation (Tables~\ref{tab:rel_cl_abs},~\ref{tab:rel_cl_paired} and~\ref{tab:rel_cl_combined}), we resample scenes and keep both rollouts of a scene together, since they are strongly correlated (intraclass correlation 0.59--0.92; $100{,}000$ resamples). For LingoQA evaluation (Tables~\ref{tab:rel_lang_abs} and~\ref{tab:rel_lang_paired}), we resample the 100 segments and keep the five questions of each segment together. For open-loop evaluation (Tables~\ref{tab:rel_ol_abs} and~\ref{tab:rel_ol_paired}) and CoC likelihood (Tables~\ref{tab:rel_lang_abs} and~\ref{tab:rel_lang_paired}), we resample clips ($200{,}000$ resamples). All methods are evaluated on the same resamples, so differences between methods are paired, and we call a difference significant when its 95\% paired interval excludes zero.

Because OOD-Val262 contains only 262 clips, we also evaluate open-loop prediction on OOD-train, the 1,271 OOD clips in the training split of the same OOD pool. These clips are disjoint from OOD-Val262 and from our calibration set but may have been seen by the base model, so this set measures how well each method retains the Baseline's performance on long-tail scenarios rather than generalization to unseen data.

These tests support three conclusions. First, neither objective is sufficient on its own. \ours{} is significantly better than CE-only on every benchmark, and significantly better than FM-only in reasoning, in long-tail (OOD) open-loop evaluation and in closed-loop evaluation over the combined scenes. Second, in open-loop evaluation, \ours{} performs on par with the strongest pruning baseline on in-distribution data (Test500), and on long-tail (OOD) data it retains the performance of the dense Baseline significantly better than every other compressed method. Third, \ours{} is the only compressed model that preserves reasoning. Its LingoQA accuracy is not significantly different from that of the dense Baseline, whereas every other compressed method loses at least 28 percentage points.

\begin{table}[!h]
\centering
\caption{\textbf{Closed-loop metrics with 95\% confidence intervals.} The best result is in \textbf{bold} and the second best is \underline{underlined}.}
\label{tab:rel_cl_abs}
\setlength{\tabcolsep}{4pt}
\renewcommand{\arraystretch}{1.25}
\resizebox{\linewidth}{!}{%
\begin{tabular}{l|c|ccc|ccc}
\toprule
\multirow{2}{*}{Method} & Expert & \multicolumn{3}{c|}{Origin150} & \multicolumn{3}{c}{Hard100} \\
 & Pruning & Scene Score\,$\uparrow$ & Offroad\,$\downarrow$ & At-fault Coll.\,$\downarrow$ & Scene Score\,$\uparrow$ & Offroad\,$\downarrow$ & At-fault Coll.\,$\downarrow$ \\
\midrule\midrule
Baseline & \xmark & \civs{0.750}{0.698, 0.800} & \civs{0.070}{0.037, 0.110} & \civs{0.040}{0.017, 0.067} & \civs{0.510}{0.435, 0.586} & \civs{0.265}{0.185, 0.350} & \civs{0.030}{0.005, 0.065} \\
\midrule
Wanda & \xmark & \civs{0.621}{0.558, 0.683} & \civs{0.093}{0.053, 0.140} & \civs{0.147}{0.097, 0.200} & \civs{0.406}{0.351, 0.461} & \civs{\textbf{0.140}}{0.080, 0.205} & \civs{0.060}{0.020, 0.105}  \\
LLM-Pruner & \xmark & \civs{\underline{0.811}}{0.761, 0.857} & \civs{\textbf{0.037}}{0.013, 0.067} & \civs{0.057}{0.027, 0.093} & \civs{0.575}{0.503, 0.646} & \civs{\underline{0.160}}{0.095, 0.230} & \civs{0.035}{0.005, 0.075}  \\
T\'yr-the-Pruner & \xmark & \civs{0.786}{0.731, 0.839} & \civs{0.087}{0.047, 0.133} & \civs{\underline{0.030}}{0.007, 0.057} & \civs{0.586}{0.508, 0.661} & \civs{0.215}{0.140, 0.295} & \civs{\textbf{0.020}}{0.000, 0.045}  \\
\midrule
FM-only & \xmark & \civs{0.783}{0.729, 0.835} & \civs{0.087}{0.047, 0.130} & \civs{0.033}{0.010, 0.060} & \civs{0.546}{0.472, 0.620} & \civs{0.260}{0.185, 0.340} & \civs{0.025}{0.005, 0.055}  \\
CE-only & \xmark & \civs{0.660}{0.603, 0.716} & \civs{0.090}{0.050, 0.133} & \civs{0.073}{0.040, 0.113} & \civs{0.416}{0.358, 0.474} & \civs{0.200}{0.130, 0.275} & \civs{0.040}{0.010, 0.075}  \\
\rowcolor{gray!15}
\textbf{\ours{} (Ours)} & \xmark & \civs{\textbf{0.828}}{0.779, 0.875} & \civs{\underline{0.067}}{0.033, 0.107} & \civs{\textbf{0.023}}{0.003, 0.047} & \civs{\underline{0.595}}{0.517, 0.672} & \civs{0.190}{0.120, 0.265} & \civs{0.040}{0.010, 0.075}  \\
\rowcolor{gray!15}
\textbf{\ours{} (Ours)} & \cmark & \civs{0.793}{0.740, 0.843} & \civs{0.083}{0.047, 0.127} & \civs{\underline{0.030}}{0.007, 0.057} & \civs{\textbf{0.600}}{0.522, 0.676} & \civs{0.200}{0.125, 0.280} & \civs{\textbf{0.020}}{0.005, 0.040}  \\
\bottomrule
\end{tabular}}
\end{table}

\begin{table}[!ht]
\centering
\caption{\textbf{Paired differences from \ours{} in closed-loop driving on each set (95\% CI).} Differences are signed so that a positive value favors \ours{}, that is, \ours{} minus the method for the scene score and the method minus \ours{} for the off-road and at-fault collision rates.}
\label{tab:rel_cl_paired}
\setlength{\tabcolsep}{4pt}
\renewcommand{\arraystretch}{1.25}
\resizebox{\linewidth}{!}{%
\begin{tabular}{l|c|ccc|ccc}
\toprule
\multirow{2}{*}{Method} & Expert & \multicolumn{3}{c|}{Origin150} & \multicolumn{3}{c}{Hard100} \\
 & Pruning & Scene Score & Offroad & At-fault Coll. & Scene Score & Offroad & At-fault Coll. \\
\midrule\midrule
Baseline & \xmark & \civs{+0.079}{+0.034, +0.123} & \civs{+0.003}{-0.027, +0.033} & \civs{+0.017}{-0.010, +0.043} & \civs{+0.085}{+0.030, +0.142} & \civs{+0.075}{+0.020, +0.135} & \civs{-0.010}{-0.040, +0.020} \\
\midrule
Wanda & \xmark & \civs{+0.207}{+0.140, +0.275} & \civs{+0.027}{-0.013, +0.067} & \civs{+0.123}{+0.070, +0.180} & \civs{+0.189}{+0.102, +0.276} & \civs{-0.050}{-0.125, +0.025} & \civs{+0.020}{-0.035, +0.075} \\
LLM-Pruner & \xmark & \civs{+0.018}{-0.029, +0.065} & \civs{-0.030}{-0.057, -0.003} & \civs{+0.033}{+0.003, +0.067} & \civs{+0.020}{-0.060, +0.099} & \civs{-0.030}{-0.105, +0.045} & \civs{-0.005}{-0.055, +0.045} \\
T\'yr-the-Pruner & \xmark & \civs{+0.042}{+0.000, +0.085} & \civs{+0.020}{-0.010, +0.053} & \civs{+0.007}{-0.017, +0.033} & \civs{+0.010}{-0.049, +0.070} & \civs{+0.025}{-0.040, +0.090} & \civs{-0.020}{-0.055, +0.015} \\
\midrule
FM-only & \xmark & \civs{+0.045}{-0.003, +0.094} & \civs{+0.020}{-0.013, +0.053} & \civs{+0.010}{-0.023, +0.043} & \civs{+0.049}{+0.003, +0.096} & \civs{+0.070}{+0.015, +0.130} & \civs{-0.015}{-0.040, +0.010} \\
CE-only & \xmark & \civs{+0.168}{+0.114, +0.223} & \civs{+0.023}{-0.010, +0.060} & \civs{+0.050}{+0.013, +0.093} & \civs{+0.180}{+0.100, +0.257} & \civs{+0.010}{-0.065, +0.085} & \civs{+0.000}{-0.035, +0.035} \\
\rowcolor{gray!15}
\textbf{\ours{} (Ours)} & \cmark & \civs{+0.035}{+0.000, +0.072} & \civs{+0.017}{-0.017, +0.050} & \civs{+0.007}{-0.017, +0.030} & \civs{-0.004}{-0.038, +0.030} & \civs{+0.010}{-0.020, +0.045} & \civs{-0.020}{-0.050, +0.005} \\
\bottomrule
\end{tabular}}
\end{table}

\begin{table}[!ht]
\centering
\caption{\textbf{Paired differences from \ours{} over the combined 250 scenes (95\% CI).} Scenes are resampled within each set. Signs follow Table~\ref{tab:rel_cl_paired}.}
\label{tab:rel_cl_combined}
\setlength{\tabcolsep}{4pt}
\renewcommand{\arraystretch}{1.15}
\begin{tabular}{l|c|ccc}
\toprule
\multirow{2}{*}{Method} & Expert & \multicolumn{3}{c}{Origin150 + Hard100} \\
 & Pruning & Scene Score & Offroad & At-fault Coll. \\
\midrule\midrule
Baseline & \xmark & \cival{+0.081}{+0.046, +0.116} & \cival{+0.032}{+0.002, +0.062} & \cival{+0.006}{-0.014, +0.026} \\
\midrule
Wanda & \xmark & \cival{+0.200}{+0.147, +0.253} & \cival{-0.004}{-0.042, +0.034} & \cival{+0.082}{+0.044, +0.122} \\
LLM-Pruner & \xmark & \cival{+0.019}{-0.024, +0.061} & \cival{-0.030}{-0.064, +0.004} & \cival{+0.018}{-0.010, +0.046} \\
T\'yr-the-Pruner & \xmark & \cival{+0.029}{-0.005, +0.064} & \cival{+0.022}{-0.010, +0.056} & \cival{-0.004}{-0.024, +0.016} \\
\midrule
FM-only & \xmark & \cival{+0.047}{+0.012, +0.081} & \cival{+0.040}{+0.010, +0.070} & \cival{+0.000}{-0.022, +0.024} \\
CE-only & \xmark & \cival{+0.173}{+0.127, +0.218} & \cival{+0.018}{-0.018, +0.054} & \cival{+0.030}{+0.004, +0.058} \\
\rowcolor{gray!15}
\textbf{\ours{} (Ours)} & \cmark & \cival{+0.020}{-0.005, +0.046} & \cival{+0.014}{-0.008, +0.038} & \cival{-0.004}{-0.022, +0.014} \\
\bottomrule
\end{tabular}
\end{table}

\begin{table}[!ht]
\centering
\caption{\textbf{Open-loop prediction at 6.4\,s with 95\% confidence intervals (m).} The best result is in \textbf{bold} and the second best is \underline{underlined}.}
\label{tab:rel_ol_abs}
\setlength{\tabcolsep}{4pt}
\renewcommand{\arraystretch}{1.25}
\resizebox{\linewidth}{!}{%
\begin{tabular}{l|c|c|cc|cc|cc}
\toprule
\multirow{2}{*}{Method} & Expert & Pruning & \multicolumn{2}{c|}{Test500} & \multicolumn{2}{c|}{OOD-Val262} & \multicolumn{2}{c}{OOD-train} \\
 & Pruning & Ratio & minADE@6 (m)\,$\downarrow$ & minFDE@6 (m)\,$\downarrow$ & minADE@6 (m)\,$\downarrow$ & minFDE@6 (m)\,$\downarrow$ & minADE@6 (m)\,$\downarrow$ & minFDE@6 (m)\,$\downarrow$ \\
\midrule\midrule
Baseline & \xmark & 0\% & \civs{0.842}{0.765, 0.922} & \civs{2.267}{2.009, 2.540} & \civs{1.000}{0.847, 1.177} & \civs{2.496}{2.041, 3.020} & \civs{0.928}{0.856, 1.007} & \civs{2.354}{2.144, 2.582} \\
\midrule
Wanda & \xmark & 24\% & \civs{2.975}{2.729, 3.233} & \civs{8.269}{7.532, 9.038} & \civs{2.790}{2.461, 3.141} & \civs{6.884}{5.983, 7.842} & \civs{2.832}{2.659, 3.013} & \civs{6.873}{6.418, 7.348}  \\
LLM-Pruner & \xmark & 24\% & \civs{1.451}{1.312, 1.600} & \civs{3.889}{3.485, 4.314} & \civs{1.662}{1.432, 1.916} & \civs{4.153}{3.525, 4.844} & \civs{1.597}{1.485, 1.716} & \civs{3.738}{3.450, 4.046}  \\
T\'yr-the-Pruner & \xmark & 24\% & \civs{\textbf{0.950}}{0.862, 1.044} & \civs{2.570}{2.289, 2.867} & \civs{1.177}{1.012, 1.366} & \civs{2.990}{2.504, 3.543} & \civs{1.178}{1.090, 1.272} & \civs{2.768}{2.536, 3.014}  \\
\midrule
FM-only & \xmark & 24\% & \civs{1.008}{0.917, 1.105} & \civs{2.720}{2.423, 3.036} & \civs{1.178}{1.011, 1.371} & \civs{2.960}{2.475, 3.515} & \civs{1.184}{1.098, 1.278} & \civs{2.897}{2.661, 3.150}  \\
CE-only & \xmark & 24\% & \civs{1.584}{1.433, 1.746} & \civs{4.500}{4.022, 5.007} & \civs{1.546}{1.328, 1.792} & \civs{4.010}{3.364, 4.733} & \civs{1.505}{1.402, 1.614} & \civs{3.737}{3.459, 4.031}  \\
\rowcolor{gray!15}
\textbf{\ours{} (Ours)} & \xmark & 24\% & \civs{\textbf{0.950}}{0.869, 1.034} & \civs{\underline{2.546}}{2.276, 2.832} & \civs{\underline{1.119}}{0.958, 1.307} & \civs{\underline{2.851}}{2.362, 3.410} & \civs{\textbf{1.038}}{0.959, 1.124} & \civs{\underline{2.541}}{2.320, 2.778}  \\
\rowcolor{gray!15}
\textbf{\ours{} (Ours)} & \cmark & 39.4\% & \civs{0.951}{0.871, 1.035} & \civs{\textbf{2.539}}{2.272, 2.822} & \civs{\textbf{1.118}}{0.957, 1.304} & \civs{\textbf{2.844}}{2.357, 3.400} & \civs{\underline{1.039}}{0.960, 1.124} & \civs{\textbf{2.537}}{2.318, 2.772}  \\
\bottomrule
\end{tabular}}
\end{table}

\begin{table}[!ht]
\centering
\caption{\textbf{Paired differences from \ours{} in open-loop prediction at 6.4\,s (method minus \ours{}, m, 95\% CI).} A positive value means a larger error than \ours{}.}
\label{tab:rel_ol_paired}
\setlength{\tabcolsep}{4pt}
\renewcommand{\arraystretch}{1.25}
\resizebox{\linewidth}{!}{%
\begin{tabular}{l|c|c|cc|cc|cc}
\toprule
\multirow{2}{*}{Method} & Expert & Pruning & \multicolumn{2}{c|}{Test500} & \multicolumn{2}{c|}{OOD-Val262} & \multicolumn{2}{c}{OOD-train} \\
 & Pruning & Ratio & minADE@6 (m) & minFDE@6 (m) & minADE@6 (m) & minFDE@6 (m) & minADE@6 (m) & minFDE@6 (m) \\
\midrule\midrule
Baseline & \xmark & 0\% & \civs{-0.108}{-0.161, -0.055} & \civs{-0.279}{-0.462, -0.097} & \civs{-0.120}{-0.226, -0.021} & \civs{-0.356}{-0.682, -0.052} & \civs{-0.111}{-0.167, -0.058} & \civs{-0.187}{-0.353, -0.030} \\
\midrule
Wanda & \xmark & 24\% & \civs{+2.026}{+1.794, +2.268} & \civs{+5.723}{+5.016, +6.457} & \civs{+1.671}{+1.380, +1.980} & \civs{+4.033}{+3.245, +4.872} & \civs{+1.794}{+1.642, +1.951} & \civs{+4.332}{+3.922, +4.758} \\
LLM-Pruner & \xmark & 24\% & \civs{+0.502}{+0.388, +0.623} & \civs{+1.343}{+1.012, +1.695} & \civs{+0.543}{+0.372, +0.729} & \civs{+1.302}{+0.851, +1.782} & \civs{+0.559}{+0.478, +0.643} & \civs{+1.197}{+0.980, +1.422} \\
T\'yr-the-Pruner & \xmark & 24\% & \civs{+0.001}{-0.055, +0.058} & \civs{+0.024}{-0.141, +0.192} & \civs{+0.057}{-0.046, +0.157} & \civs{+0.138}{-0.129, +0.391} & \civs{+0.139}{+0.082, +0.197} & \civs{+0.227}{+0.071, +0.381} \\
\midrule
FM-only & \xmark & 24\% & \civs{+0.058}{+0.001, +0.118} & \civs{+0.174}{-0.004, +0.365} & \civs{+0.059}{-0.035, +0.148} & \civs{+0.108}{-0.134, +0.339} & \civs{+0.146}{+0.099, +0.192} & \civs{+0.356}{+0.231, +0.481} \\
CE-only & \xmark & 24\% & \civs{+0.634}{+0.509, +0.768} & \civs{+1.954}{+1.555, +2.381} & \civs{+0.427}{+0.277, +0.596} & \civs{+1.158}{+0.740, +1.639} & \civs{+0.467}{+0.392, +0.545} & \civs{+1.196}{+0.989, +1.412} \\
\rowcolor{gray!15}
\textbf{\ours{} (Ours)} & \cmark & 39.4\% & \civs{+0.001}{-0.002, +0.004} & \civs{-0.007}{-0.018, +0.003} & \civs{-0.002}{-0.007, +0.003} & \civs{-0.007}{-0.027, +0.012} & \civs{+0.000}{-0.002, +0.003} & \civs{-0.004}{-0.013, +0.005} \\
\bottomrule
\end{tabular}}
\end{table}

\begin{table}[!ht]
\centering
\caption{\textbf{LingoQA accuracy and CoC NLL with 95\% confidence intervals.} The NLL is computed on the CoC generated by the Baseline for Test500 and on the ground-truth CoC for OOD-train. The best result is in \textbf{bold} and the second best is \underline{underlined}.}
\label{tab:rel_lang_abs}
\setlength{\tabcolsep}{4pt}
\renewcommand{\arraystretch}{1.15}
\begin{tabular}{l|c|cc}
\toprule
\multirow{2}{*}{Method} & \multirow{2}{*}{LingoQA (\%)\,$\uparrow$} & \multicolumn{2}{c}{CoC NLL\,$\downarrow$} \\
 & & Test500 & OOD-train \\
\midrule\midrule
Baseline & \cival{73.2}{69.4, 76.8} & \cival{0.175}{0.161, 0.190} & \cival{2.977}{2.918, 3.037} \\
\midrule
Wanda & \cival{9.2}{6.8, 11.8} & \cival{2.183}{2.036, 2.334} & \cival{4.756}{4.642, 4.873}  \\
LLM-Pruner & \cival{\underline{44.8}}{40.0, 49.6} & \cival{0.462}{0.419, 0.505} & \cival{3.215}{3.152, 3.278}  \\
T\'yr-the-Pruner & \cival{34.2}{30.0, 38.6} & \cival{0.516}{0.468, 0.568} & \cival{3.108}{3.045, 3.171}  \\
\midrule
FM-only & \cival{37.0}{32.4, 41.8} & \cival{0.586}{0.540, 0.634} & \cival{\underline{3.106}}{3.045, 3.168}  \\
CE-only & \cival{30.2}{25.6, 34.8} & \cival{\underline{0.313}}{0.286, 0.340} & \cival{3.152}{3.087, 3.219}  \\
\rowcolor{gray!15}
\textbf{\ours{} (Ours)} & \cival{\textbf{68.8}}{64.4, 73.0} & \cival{\textbf{0.252}}{0.228, 0.278} & \cival{\textbf{3.007}}{2.945, 3.070}  \\
\midrule
Always answering ``yes'' & 37.6 & -- & -- \\
\bottomrule
\end{tabular}
\end{table}

\begin{table}[!ht]
\centering
\caption{\textbf{Paired differences on LingoQA and CoC NLL (95\% CI).} For LingoQA, $\Delta$ is \ours{} minus the method in percentage points, so a positive value favors \ours{}, and the last LingoQA column is the method minus always answering ``yes''. For NLL, $\Delta$ is the method minus \ours{}, so a positive value also favors \ours{}.}
\label{tab:rel_lang_paired}
\setlength{\tabcolsep}{4pt}
\renewcommand{\arraystretch}{1.15}
\resizebox{\linewidth}{!}{%
\begin{tabular}{l|cc|cc}
\toprule
\multirow{2}{*}{Method} & \multicolumn{2}{c|}{LingoQA (pp)} & \multicolumn{2}{c}{CoC NLL} \\
 & $\Delta$ vs.\ \ours{} & vs.\ always ``yes'' & Test500 & OOD-train \\
\midrule\midrule
Baseline & \cival{-4.4}{-9.4, +0.4} & \cival{+35.6}{+29.6, +41.4} & \cival{-0.077}{-0.091, -0.063} & \cival{-0.030}{-0.057, -0.003} \\
\midrule
Wanda & \cival{+59.6}{+55.0, +64.2} & \cival{-28.4}{-33.8, -23.2} & \cival{+1.931}{+1.792, +2.074} & \cival{+1.749}{+1.662, +1.838} \\
LLM-Pruner & \cival{+24.0}{+18.6, +29.4} & \cival{+7.2}{+0.2, +14.0} & \cival{+0.209}{+0.178, +0.243} & \cival{+0.208}{+0.186, +0.230} \\
T\'yr-the-Pruner & \cival{+34.6}{+28.6, +40.4} & \cival{-3.4}{-10.2, +3.6} & \cival{+0.264}{+0.235, +0.295} & \cival{+0.101}{+0.080, +0.122} \\
\midrule
FM-only & \cival{+31.8}{+26.4, +37.2} & \cival{-0.6}{-7.2, +6.0} & \cival{+0.333}{+0.306, +0.362} & \cival{+0.099}{+0.075, +0.123} \\
CE-only & \cival{+38.6}{+33.2, +44.0} & \cival{-7.4}{-14.2, -0.6} & \cival{+0.060}{+0.046, +0.076} & \cival{+0.145}{+0.127, +0.164} \\
\rowcolor{gray!15}
\textbf{\ours{} (Ours)} & -- & \cival{+31.2}{+24.6, +37.8} & -- & -- \\
\bottomrule
\end{tabular}}
\end{table}

\clearpage
\subsection{Robustness at a Higher Pruning Ratio}
\label{app:higher_ratio}

To test whether the differences between methods grow under stronger compression, we prune every method at a VLM pruning ratio of $r = 0.5$, which removes $30.0\%$ of all parameters. We evaluate closed-loop driving on Origin150 and Hard100, open-loop prediction and LingoQA with the bootstrap of Appendix~\ref{app:reliability}, and call a difference significant when its paired 95\% interval excludes zero.

\paragraph{Closed-loop driving.}
As shown in Tables~\ref{tab:hr_cl_abs} and~\ref{tab:hr_cl_paired}, \ours{} achieves the highest scene score on both sets. On Origin150, it is significantly better than every other compressed method in scene score and at-fault collisions. On Hard100, its scene score is significantly higher than those of all other compressed methods except T\'yr-the-Pruner, whose difference ($+0.037$ $[-0.037, +0.111]$) is not significant. At $r = 0.4$, \ours{} already achieves the highest Origin150 scene score, but its margins are small, as shown in Table~\ref{tab:rel_cl_paired} (0.042 over T\'yr-the-Pruner and 0.018 over LLM-Pruner). At $r = 0.5$, however, these margins grow to 0.124 and 0.232 on Origin150 and are clearly significant.

\paragraph{Open-loop prediction.}
As shown in Table~\ref{tab:hr_ol_abs}, \ours{} is significantly better than every other method in open-loop prediction. The closest method, T\'yr-the-Pruner, has a minADE@6 higher by 0.813\,m on Test500 and 0.749\,m on OOD-Val262. Every compressed method is significantly worse than the dense Baseline at this ratio, but the gap of \ours{} on Test500 (0.395\,m) is about one third of that of T\'yr-the-Pruner (1.208\,m). The ranking in open-loop prediction closely follows the ranking in closed-loop driving.

\paragraph{LingoQA and CoC likelihood.}
Table~\ref{tab:hr_lang_abs} reports LingoQA accuracy and CoC NLL. \ours{} keeps 47.4\% LingoQA accuracy, significantly above all other methods and 13.2 percentage points above LLM-Pruner, the next best, which is not distinguishable from always answering ``yes''. CE-only falls significantly below always answering ``yes'', by 10.6 percentage points, and T\'yr-the-Pruner, FM-only and Wanda fall to 1.6--7.2\%. \ours{} also has the lowest CoC NLL among the compressed models (0.376), below CE-only by 0.164.

At $r = 0.5$, \ours{} is therefore significantly better than every other compressed method in open-loop prediction and in closed-loop driving on Origin150, achieves the highest closed-loop scene score on both sets, and is the only method that preserves reasoning. It degrades less under more aggressive pruning while retaining the reasoning ability of the dense model. 

\begin{table}[!h]
\centering
\caption{\textbf{Closed-loop driving at $r = 0.5$, with 95\% confidence intervals.} The Baseline is the same as in Table~\ref{tab:rel_cl_abs}. The best result is in \textbf{bold} and the second best is \underline{underlined}.}
\label{tab:hr_cl_abs}
\setlength{\tabcolsep}{4pt}
\renewcommand{\arraystretch}{1.25}
\resizebox{\linewidth}{!}{%
\begin{tabular}{l|c|c|ccc|ccc}
\toprule
\multirow{2}{*}{Method} & Expert & Pruning & \multicolumn{3}{c|}{Origin150} & \multicolumn{3}{c}{Hard100} \\
 & Pruning & Ratio & Scene Score\,$\uparrow$ & Offroad\,$\downarrow$ & At-fault Coll.\,$\downarrow$ & Scene Score\,$\uparrow$ & Offroad\,$\downarrow$ & At-fault Coll.\,$\downarrow$ \\
\midrule\midrule
Baseline & \xmark & 0\% & \civs{0.750}{0.698, 0.800} & \civs{0.070}{0.037, 0.110} & \civs{0.040}{0.017, 0.067} & \civs{0.510}{0.435, 0.586} & \civs{0.265}{0.185, 0.350} & \civs{0.030}{0.005, 0.065} \\
\midrule
Wanda & \xmark & 30\% & \civs{0.350}{0.291, 0.410} & \civs{0.280}{0.213, 0.350} & \civs{0.180}{0.123, 0.240} & \civs{0.266}{0.213, 0.322} & \civs{0.290}{0.205, 0.380} & \civs{0.025}{0.005, 0.055} \\
LLM-Pruner & \xmark & 30\% & \civs{0.634}{0.570, 0.697} & \civs{\underline{0.073}}{0.037, 0.117} & \civs{0.143}{0.093, 0.200} & \civs{0.454}{0.383, 0.525} & \civs{0.210}{0.135, 0.290} & \civs{0.055}{0.020, 0.095} \\
T\'yr-the-Pruner & \xmark & 30\% & \civs{\underline{0.742}}{0.682, 0.800} & \civs{\underline{0.073}}{0.037, 0.117} & \civs{\underline{0.087}}{0.047, 0.130} & \civs{\underline{0.614}}{0.539, 0.687} & \civs{\textbf{0.140}}{0.080, 0.210} & \civs{\underline{0.020}}{0.000, 0.045} \\
\midrule
FM-only & \xmark & 30\% & \civs{0.551}{0.494, 0.607} & \civs{0.103}{0.060, 0.153} & \civs{0.097}{0.057, 0.143} & \civs{0.289}{0.241, 0.339} & \civs{0.240}{0.165, 0.320} & \civs{0.070}{0.030, 0.115} \\
CE-only & \xmark & 30\% & \civs{0.279}{0.239, 0.322} & \civs{0.170}{0.117, 0.230} & \civs{0.140}{0.090, 0.197} & \civs{0.153}{0.120, 0.188} & \civs{0.345}{0.260, 0.435} & \civs{0.070}{0.030, 0.120} \\
\rowcolor{gray!15}
\textbf{\ours{} (Ours)} & \xmark & 30\% & \civs{\textbf{0.866}}{0.821, 0.908} & \civs{\textbf{0.047}}{0.020, 0.080} & \civs{\textbf{0.037}}{0.013, 0.063} & \civs{\textbf{0.651}}{0.574, 0.725} & \civs{\underline{0.195}}{0.125, 0.270} & \civs{\textbf{0.010}}{0.000, 0.025} \\
\bottomrule
\end{tabular}}
\end{table}

\begin{table}[!h]
\centering
\caption{\textbf{Paired differences from \ours{} in closed-loop driving at $r = 0.5$ (95\% CI).} Differences are signed so that a positive value favors \ours{}, that is, \ours{} minus the method for the scene score and the method minus \ours{} for the off-road and at-fault collision rates.}
\label{tab:hr_cl_paired}
\setlength{\tabcolsep}{4pt}
\renewcommand{\arraystretch}{1.25}
\resizebox{\linewidth}{!}{%
\begin{tabular}{l|c|c|ccc|ccc}
\toprule
\multirow{2}{*}{Method} & Expert & Pruning & \multicolumn{3}{c|}{Origin150} & \multicolumn{3}{c}{Hard100} \\
 & Pruning & Ratio & Scene Score & Offroad & At-fault Coll. & Scene Score & Offroad & At-fault Coll. \\
\midrule\midrule
Baseline & \xmark & 0\% & \civs{+0.116}{+0.072, +0.162} & \civs{+0.023}{-0.003, +0.053} & \civs{+0.003}{-0.027, +0.033} & \civs{+0.140}{+0.070, +0.212} & \civs{+0.070}{-0.005, +0.145} & \civs{+0.020}{-0.005, +0.050} \\
\midrule
Wanda & \xmark & 30\% & \civs{+0.516}{+0.448, +0.584} & \civs{+0.233}{+0.163, +0.303} & \civs{+0.143}{+0.080, +0.210} & \civs{+0.385}{+0.298, +0.470} & \civs{+0.095}{-0.005, +0.195} & \civs{+0.015}{-0.010, +0.045} \\
LLM-Pruner & \xmark & 30\% & \civs{+0.232}{+0.164, +0.301} & \civs{+0.027}{-0.017, +0.070} & \civs{+0.107}{+0.050, +0.163} & \civs{+0.197}{+0.112, +0.282} & \civs{+0.015}{-0.065, +0.095} & \civs{+0.045}{+0.005, +0.090} \\
T\'yr-the-Pruner & \xmark & 30\% & \civs{+0.124}{+0.067, +0.183} & \civs{+0.027}{-0.003, +0.057} & \civs{+0.050}{+0.007, +0.097} & \civs{+0.037}{-0.037, +0.111} & \civs{-0.055}{-0.115, +0.005} & \civs{+0.010}{-0.015, +0.040} \\
\midrule
FM-only & \xmark & 30\% & \civs{+0.315}{+0.259, +0.372} & \civs{+0.057}{+0.010, +0.107} & \civs{+0.060}{+0.017, +0.107} & \civs{+0.362}{+0.277, +0.445} & \civs{+0.045}{-0.050, +0.140} & \civs{+0.060}{+0.015, +0.110} \\
CE-only & \xmark & 30\% & \civs{+0.587}{+0.531, +0.640} & \civs{+0.123}{+0.067, +0.183} & \civs{+0.103}{+0.043, +0.163} & \civs{+0.498}{+0.421, +0.573} & \civs{+0.150}{+0.055, +0.245} & \civs{+0.060}{+0.015, +0.110} \\
\bottomrule
\end{tabular}}
\end{table}

\begin{table}[!h]
\centering
\caption{\textbf{Open-loop prediction at 6.4\,s at $r = 0.5$, with 95\% confidence intervals (m).} The Baseline is the same as in Table~\ref{tab:rel_ol_abs}. The best result is in \textbf{bold} and the second best is \underline{underlined}.}
\label{tab:hr_ol_abs}
\renewcommand{\arraystretch}{1.15}
\resizebox{\linewidth}{!}{%
\begin{tabular}{l|c|c|cc|cc}
\toprule
\multirow{2}{*}{Method} & Expert & Pruning & \multicolumn{2}{c|}{Test500} & \multicolumn{2}{c}{OOD-val262} \\
 & Pruning & Ratio & minADE@6 (m)\,$\downarrow$ & minFDE@6 (m)\,$\downarrow$ & minADE@6 (m)\,$\downarrow$ & minFDE@6 (m)\,$\downarrow$ \\
\midrule\midrule
Baseline         & \xmark & 0\%  & \cival{0.842}{0.765, 0.922} & \cival{2.267}{2.009, 2.540} & \cival{1.000}{0.847, 1.177} & \cival{2.496}{2.041, 3.020} \\
\midrule
Wanda & \xmark & 30\% & \cival{2.901}{2.658, 3.157} & \cival{7.869}{7.158, 8.612} & \cival{2.761}{2.439, 3.103} & \cival{6.917}{6.028, 7.861}  \\
LLM-Pruner & \xmark & 30\% & \cival{2.533}{2.312, 2.766} & \cival{6.662}{6.040, 7.319} & \cival{2.628}{2.288, 2.993} & \cival{6.613}{5.692, 7.600}  \\
T\'yr-the-Pruner & \xmark & 30\% & \cival{\underline{2.050}}{1.854, 2.257} & \cival{\underline{5.339}}{4.788, 5.921} & \cival{\underline{2.165}}{1.905, 2.443} & \cival{\underline{5.202}}{4.525, 5.917}  \\
\midrule
FM-only & \xmark & 30\% & \cival{2.822}{2.568, 3.094} & \cival{8.278}{7.505, 9.102} & \cival{2.436}{2.111, 2.788} & \cival{6.753}{5.789, 7.799}  \\
CE-only & \xmark & 30\% & \cival{5.151}{4.761, 5.560} & \cival{15.192}{14.014, 16.432} & \cival{3.899}{3.468, 4.358} & \cival{11.230}{9.930, 12.603}  \\
\rowcolor{gray!15}
\textbf{\ours{} (Ours)} & \xmark & 30\% & \cival{\textbf{1.237}}{1.126, 1.356} & \cival{\textbf{3.338}}{2.980, 3.720} & \cival{\textbf{1.416}}{1.217, 1.640} & \cival{\textbf{3.616}}{3.027, 4.266}  \\
\bottomrule
\end{tabular}}
\end{table}

\begin{table}[!h]
\centering
\caption{\textbf{LingoQA accuracy and CoC NLL at $r = 0.5$, with 95\% confidence intervals.} The NLL is computed on the CoC generated by the Baseline for Test500. The Baseline is the same as in Table~\ref{tab:rel_lang_abs}. The best result is in \textbf{bold} and the second best is \underline{underlined}.}
\label{tab:hr_lang_abs}
\setlength{\tabcolsep}{4pt}
\renewcommand{\arraystretch}{1.15}
\begin{tabular}{l|c|c}
\toprule
Method & LingoQA (\%)\,$\uparrow$ & CoC NLL, Test500\,$\downarrow$ \\
\midrule\midrule
Baseline & \cival{73.2}{69.4, 76.8} & \cival{0.175}{0.161, 0.190} \\
\midrule
Wanda & \cival{7.2}{5.0, 9.6} & \cival{8.344}{7.985, 8.712}  \\
LLM-Pruner & \cival{\underline{34.2}}{29.6, 39.0} & \cival{0.976}{0.923, 1.031}  \\
T\'yr-the-Pruner & \cival{6.2}{3.8, 8.8} & \cival{2.254}{2.108, 2.408}  \\
\midrule
FM-only & \cival{1.6}{0.6, 3.0} & \cival{1.852}{1.753, 1.955}  \\
CE-only & \cival{27.0}{22.8, 31.4} & \cival{\underline{0.540}}{0.500, 0.582}  \\
\rowcolor{gray!15}
\textbf{\ours{} (Ours)} & \cival{\textbf{47.4}}{42.8, 52.0} & \cival{\textbf{0.376}}{0.337, 0.417}  \\
\midrule
Always answering ``yes'' & 37.6 & -- \\
\bottomrule
\end{tabular}
\end{table}

\clearpage

\section{Additional Experiments and Discussion}
\label{app:discussion}

\subsection{Extended Analysis}
\label{app:analysis_expansion}

\paragraph{Non-uniform layer-wise allocation based on rank correlation.}
In the main experiments, we apply the same pruning ratio to every layer so that the comparison isolates the effect of the pruning criterion. Our analysis, however, also suggests a non-uniform allocation. We base it on the CE--FM rank correlation computed on generated CoC tokens rather than on all tokens (Figure~\ref{fig:vlm_rank_correlation}). Although the all-token correlation is high in the early layers, the CoC-token correlation remains low across much of the backbone, so units in the early and middle layers can also serve CoC generation and action conditioning differently. Let $\rho_\ell$ be the CoC-token rank correlation of layer $\ell$, and let $\bar\rho$ be its mean over the layers whose ratio is adjusted. We set the pruning ratio of layer $\ell$ to $0.4 + a\,(\rho_\ell - \bar\rho)$, so that layers with a lower correlation are pruned less.

Two adjustments keep this allocation stable. First, the CoC-token correlation fluctuates considerably between adjacent layers, which would cause abrupt changes in the pruning ratio between neighboring layers. Since such changes degrade performance~\citep{yin2025outlierweighedlayerwisesparsity}, we scale the deviation by $a = 0.25$, which limits it to at most 0.12 for query heads and 0.10 for MLP channels. Second, since the early layers are particularly sensitive to pruning~\citep{ma2023llm}, we keep the first $i = 4$ layers at the uniform ratio of 0.4. Query heads and MLP channels are allocated separately, units within each layer are still selected by the max-rank rule, and the total parameter reduction remains 39.4\%, as in \ours{}.

Under the same budget, this allocation lowers minADE@6 from 0.951\,m to 0.939\,m on Test500 and from 1.118\,m to 1.086\,m on OOD-Val262 (Table~\ref{tab:abl_budget_abs}). A more careful search over $a$ and $i$ may improve performance further. Our analysis can therefore guide not only the selection of units within each layer but also the pruning ratio of each layer.

\begin{table}[!h]
\centering
\caption{\textbf{Non-uniform layer-wise allocation with expert pruning (39.4\% total), with 95\% confidence intervals.}}
\label{tab:abl_budget_abs}
\setlength{\tabcolsep}{4pt}
\renewcommand{\arraystretch}{1.15}
\resizebox{\linewidth}{!}{%
\begin{tabular}{l|cc|cc}
\toprule
\multirow{2}{*}{Budget allocation} & \multicolumn{2}{c|}{Test500} & \multicolumn{2}{c}{OOD-Val262} \\
 & minADE@6 (m)\,$\downarrow$ & minFDE@6 (m)\,$\downarrow$ & minADE@6 (m)\,$\downarrow$ & minFDE@6 (m)\,$\downarrow$ \\
\midrule\midrule
Uniform & \cival{0.951}{0.870, 1.035} & \cival{2.539}{2.275, 2.819} & \cival{1.118}{0.955, 1.303} & \cival{2.844}{2.341, 3.397} \\
Non-uniform & \cival{0.939}{0.856, 1.027} & \cival{2.498}{2.233, 2.786} & \cival{1.086}{0.928, 1.268} & \cival{2.759}{2.268, 3.310} \\
\midrule
Difference (non-uniform $-$ uniform) & \cival{-0.012}{-0.034, +0.011} & \cival{-0.041}{-0.116, +0.038} & \cival{-0.031}{-0.076, +0.011} & \cival{-0.085}{-0.204, +0.031} \\
\bottomrule
\end{tabular}}
\end{table}

\paragraph{Step-wise expert scoring.}
We prune 93.75\% of the expert MLP channels using the importance from a single denoising step $s$ instead of the step-averaged importance (Table~\ref{tab:abl_step_abs}). Every single-step mask yields nearly the same open-loop errors as the step-averaged mask. The differences are within 0.005\,m in minADE@6 and 0.025\,m in minFDE@6 for every step. This matches the stable MLP-channel rankings in Section~\ref{sec:expert_analysis} and shows that the expert can also be scored at a single step.

\begin{table}[!h]
\centering
\caption{\textbf{Single-step and step-averaged expert scoring at 93.75\% expert MLP removal, with 95\% confidence intervals.} $t$ is the flow-matching time of step $s$. Intervals are from a clip-level bootstrap with 10,000 resamples.}
\label{tab:abl_step_abs}
\setlength{\tabcolsep}{4pt}
\renewcommand{\arraystretch}{1.15}
\resizebox{\linewidth}{!}{%
\begin{tabular}{l|c|cc|cc}
\toprule
\multirow{2}{*}{Step $s$} & \multirow{2}{*}{$t$} & \multicolumn{2}{c|}{Test500} & \multicolumn{2}{c}{OOD-Val262} \\
 & & minADE@6 (m)\,$\downarrow$ & minFDE@6 (m)\,$\downarrow$ & minADE@6 (m)\,$\downarrow$ & minFDE@6 (m)\,$\downarrow$ \\
\midrule\midrule
0 & 0.05 & \cival{0.950}{0.870, 1.035} & \cival{2.537}{2.273, 2.817} & \cival{1.117}{0.955, 1.303} & \cival{2.844}{2.342, 3.397} \\
1 & 0.15 & \cival{0.950}{0.869, 1.035} & \cival{2.537}{2.273, 2.816} & \cival{1.117}{0.955, 1.303} & \cival{2.844}{2.342, 3.397} \\
2 & 0.25 & \cival{0.950}{0.870, 1.035} & \cival{2.538}{2.274, 2.818} & \cival{1.118}{0.955, 1.303} & \cival{2.845}{2.342, 3.397} \\
3 & 0.35 & \cival{0.951}{0.870, 1.036} & \cival{2.539}{2.275, 2.819} & \cival{1.118}{0.956, 1.304} & \cival{2.846}{2.342, 3.399} \\
4 & 0.45 & \cival{0.951}{0.870, 1.036} & \cival{2.539}{2.274, 2.820} & \cival{1.118}{0.955, 1.304} & \cival{2.845}{2.342, 3.398} \\
5 & 0.55 & \cival{0.951}{0.870, 1.036} & \cival{2.540}{2.276, 2.820} & \cival{1.118}{0.955, 1.303} & \cival{2.844}{2.341, 3.397} \\
6 & 0.65 & \cival{0.950}{0.870, 1.035} & \cival{2.536}{2.273, 2.817} & \cival{1.118}{0.955, 1.303} & \cival{2.845}{2.343, 3.397} \\
7 & 0.75 & \cival{0.950}{0.870, 1.035} & \cival{2.539}{2.275, 2.819} & \cival{1.118}{0.955, 1.303} & \cival{2.845}{2.342, 3.397} \\
8 & 0.85 & \cival{0.950}{0.870, 1.035} & \cival{2.538}{2.274, 2.819} & \cival{1.118}{0.955, 1.303} & \cival{2.844}{2.342, 3.396} \\
9 & 0.95 & \cival{0.946}{0.867, 1.030} & \cival{2.514}{2.253, 2.793} & \cival{1.117}{0.954, 1.303} & \cival{2.832}{2.332, 3.386} \\
\midrule
\rowcolor{gray!15}
\textbf{Step-averaged (\ours{})} & -- & \cival{0.951}{0.870, 1.035} & \cival{2.539}{2.275, 2.819} & \cival{1.118}{0.955, 1.303} & \cival{2.844}{2.341, 3.397} \\
\bottomrule
\end{tabular}}
\end{table}

\paragraph{Expert channel selection.}
We compare our step-averaged FM Taylor score with weight-magnitude selection and random selection (three seeds) at the same expert MLP removal ratios, on top of the same \ours{}-pruned VLM (Table~\ref{tab:abl_expert_abs}). Random selection collapses at every ratio and already reaches 2.663\,m minADE@6 on Test500 at 75\% removal, close to the 3.373\,m obtained by removing all expert MLP channels. The robustness of the expert to pruning therefore depends on retaining the right channels and is not explained by over-parameterization alone. Weight magnitude performs similarly to our score on Test500 and on OOD-train, but it is significantly worse on OOD-Val262 at 87.5\% and 93.75\% removal, by 0.010\,m and 0.024\,m in minADE@6.

\begin{table}[!h]
\centering
\caption{\textbf{Expert channel selection with 95\% confidence intervals.} The first column gives the fraction of expert MLP channels removed and, below it, the total parameter reduction. Ours is the step-averaged FM Taylor score. Random selection reports the mean over three seeds with the standard deviation across seeds in parentheses, and its interval is computed on the per-clip mean over seeds. Intervals are from a clip-level bootstrap with 10,000 resamples.}
\label{tab:abl_expert_abs}
\setlength{\tabcolsep}{4pt}
\renewcommand{\arraystretch}{1.25}
\resizebox{\linewidth}{!}{%
\begin{tabular}{l|l|cc|cc|cc}
\toprule
\multirow{2}{*}{Removed} & \multirow{2}{*}{Selection} & \multicolumn{2}{c|}{Test500} & \multicolumn{2}{c|}{OOD-Val262} & \multicolumn{2}{c}{OOD-train} \\
 & & minADE@6 (m)\,$\downarrow$ & minFDE@6 (m)\,$\downarrow$ & minADE@6 (m)\,$\downarrow$ & minFDE@6 (m)\,$\downarrow$ & minADE@6 (m)\,$\downarrow$ & minFDE@6 (m)\,$\downarrow$ \\
\midrule\midrule
 & Random & \civssd{2.663}{0.055}{2.439, 2.896} & \civssd{7.037}{0.214}{6.423, 7.677} & \civssd{2.265}{0.031}{2.024, 2.526} & \civssd{6.011}{0.180}{5.316, 6.751} & \civssd{2.267}{0.101}{2.143, 2.395} & \civssd{6.070}{0.376}{5.727, 6.425} \\
 & Magnitude & \civs{0.949}{0.869, 1.034} & \civs{2.540}{2.275, 2.822} & \civs{1.119}{0.956, 1.306} & \civs{2.852}{2.349, 3.408} & \civs{1.025}{0.951, 1.107} & \civs{2.506}{2.298, 2.736} \\
 \rowcolor{gray!15}
\twoline{75\%}{(36.3\%)} & \textbf{Ours} & \civs{0.950}{0.869, 1.035} & \civs{2.543}{2.277, 2.825} & \civs{1.118}{0.955, 1.304} & \civs{2.848}{2.347, 3.403} & \civs{1.025}{0.951, 1.107} & \civs{2.506}{2.297, 2.735} \\
\midrule
 & Random & \civssd{3.088}{0.042}{2.824, 3.360} & \civssd{8.249}{0.086}{7.485, 9.031} & \civssd{2.544}{0.034}{2.269, 2.852} & \civssd{6.852}{0.204}{6.043, 7.755} & \civssd{2.566}{0.086}{2.423, 2.715} & \civssd{6.900}{0.211}{6.491, 7.320} \\
 & Magnitude & \civs{0.950}{0.870, 1.036} & \civs{2.537}{2.272, 2.822} & \civs{1.128}{0.964, 1.316} & \civs{2.881}{2.380, 3.442} & \civs{1.027}{0.953, 1.109} & \civs{2.505}{2.297, 2.735} \\
 \rowcolor{gray!15}
\twoline{87.5\%}{(38.4\%)} & \textbf{Ours} & \civs{0.950}{0.869, 1.035} & \civs{2.542}{2.277, 2.823} & \civs{1.118}{0.955, 1.304} & \civs{2.846}{2.343, 3.401} & \civs{1.025}{0.951, 1.107} & \civs{2.505}{2.297, 2.734} \\
\midrule
 & Random & \civssd{3.466}{0.228}{3.142, 3.803} & \civssd{9.273}{0.666}{8.334, 10.243} & \civssd{2.669}{0.080}{2.358, 3.023} & \civssd{7.136}{0.260}{6.244, 8.170} & \civssd{2.707}{0.102}{2.549, 2.873} & \civssd{7.234}{0.328}{6.791, 7.696} \\
 & Magnitude & \civs{0.944}{0.863, 1.030} & \civs{2.532}{2.263, 2.818} & \civs{1.141}{0.976, 1.329} & \civs{2.898}{2.401, 3.458} & \civs{1.027}{0.953, 1.109} & \civs{2.505}{2.296, 2.734} \\
 \rowcolor{gray!15}
\twoline{93.75\%}{(39.4\%)} & \textbf{Ours} & \civs{0.951}{0.870, 1.035} & \civs{2.539}{2.275, 2.819} & \civs{1.118}{0.955, 1.303} & \civs{2.844}{2.341, 3.397} & \civs{1.026}{0.952, 1.108} & \civs{2.504}{2.296, 2.732} \\
\midrule
\twoline{100\%}{(40.5\%)} & All removed & \civs{3.373}{3.078, 3.678} & \civs{8.956}{8.101, 9.822} & \civs{2.727}{2.379, 3.159} & \civs{7.246}{6.290, 8.417} & \civs{2.743}{2.577, 2.920} & \civs{7.236}{6.768, 7.736} \\
\bottomrule
\end{tabular}}
\end{table}

\clearpage
\FloatBarrier
\subsection{Ablation of the Combination Rule}
\label{app:ablations}

The max-rank rule does not weight the two objectives explicitly. This choice is deliberate, because the CE and FM importance scores have different scales: FM scores are on average about 1.6$\times$ larger than CE scores. Since the max-rank rule operates on within-layer ranks, it is unaffected by this difference in scale, and it retains a unit if the unit ranks high under either objective. It also adapts to the rank correlation of each layer without a layer-specific weight. Where the two rankings agree, the combined score closely follows the shared ranking, and where they diverge, the top-ranked units under each objective are retained.

To test whether the raw scores can be combined directly, we compare max-rank with max-raw, which takes the maximum of the raw CE and FM importance scores and thus implicitly favors FM in proportion to its larger scale (Table~\ref{tab:abl_comb_abs}). Max-raw gives a minADE@6 of 0.974\,m on Test500 and 1.173\,m on OOD-Val262 and a LingoQA accuracy of 59.2\%, compared with 0.950\,m, 1.119\,m and 68.8\% for max-rank.

We also compare max-rank with the mean of the two ranks and with the mean of the raw importance scores. The two rank-based rules perform similarly, and their paired differences are not significant on any metric, whereas both rules that combine raw scores degrade reasoning. Letting the scale of one objective determine its priority is therefore harmful, which supports combining the objectives through their within-layer ranks.

\begin{table}[!h]
\centering
\caption{\textbf{Combination rules with 95\% confidence intervals.} All rules remove the same number of query heads and MLP channels in every layer. Intervals are from segment-level (LingoQA) and clip-level (open-loop) bootstraps with 200,000 resamples. The best result is in \textbf{bold} and the second best is \underline{underlined}.}
\label{tab:abl_comb_abs}
\setlength{\tabcolsep}{4pt}
\renewcommand{\arraystretch}{1.15}
\resizebox{\linewidth}{!}{%
\begin{tabular}{l|c|cc|cc}
\toprule
\multirow{2}{*}{Rule} & \multirow{2}{*}{LingoQA (\%)\,$\uparrow$} & \multicolumn{2}{c|}{Test500} & \multicolumn{2}{c}{OOD-Val262} \\
 & & minADE@6 (m)\,$\downarrow$ & minFDE@6 (m)\,$\downarrow$ & minADE@6 (m)\,$\downarrow$ & minFDE@6 (m)\,$\downarrow$ \\
\midrule\midrule
Mean-raw & \cival{14.0}{10.8, 17.4} & \cival{1.216}{1.102, 1.337} & \cival{3.248}{2.896, 3.619} & \cival{1.373}{1.167, 1.604} & \cival{3.533}{2.903, 4.232}  \\
Mean-rank & \cival{\underline{67.0}}{62.6, 71.4} & \cival{\underline{0.960}}{0.874, 1.051} & \cival{\underline{2.607}}{2.322, 2.907} & \cival{\underline{1.160}}{0.991, 1.355} & \cival{2.967}{2.454, 3.546}  \\
\midrule
Max-raw & \cival{59.2}{54.6, 63.6} & \cival{0.974}{0.889, 1.065} & \cival{2.629}{2.345, 2.930} & \cival{1.173}{1.004, 1.367} & \cival{\underline{2.961}}{2.456, 3.534}  \\
\rowcolor{gray!15}
\textbf{Max-rank (\ours{})} & \cival{\textbf{68.8}}{64.4, 73.0} & \cival{\textbf{0.950}}{0.869, 1.034} & \cival{\textbf{2.546}}{2.276, 2.832} & \cival{\textbf{1.119}}{0.958, 1.307} & \cival{\textbf{2.851}}{2.362, 3.410}  \\
\bottomrule
\end{tabular}}
\end{table}

\FloatBarrier
\subsection{Generalization to Alpamayo 2 Edge}
\label{app:generalization}

To test whether the two findings that motivate \ours{} hold beyond Alpamayo~1.5, we repeat both analyses of Section~\ref{sec:vlm_analysis} and Section~\ref{sec:expert_analysis} on Alpamayo 2 Edge~\citep{nvidia2026alpamayo2edge}, a compact model of the Alpamayo 2 family with 2.65B parameters. Its VLM pairs a SigLIP2 vision encoder with a 28-layer language model from the Cosmos 3 Edge backbone (16 query heads and 9,216 MLP channels per layer), and its 28-layer action expert (0.2B parameters) reads the VLM KV cache layer by layer and generates trajectories with ten flow-matching steps, as in Alpamayo~1.5. We use the same Calib100 clips, importance protocol and prompt as for Alpamayo~1.5. Both findings largely hold.

\begin{figure}[p]
    \centering
    \begin{subfigure}[t]{0.32\linewidth}
        \centering
        \includegraphics[width=\linewidth]{fig/main/vlm_rank_correlation.pdf}
        \caption{Rank correlation}
        \label{fig:a15_vlm_rank}
    \end{subfigure}
    \hfill
    \begin{subfigure}[t]{0.32\linewidth}
        \centering
        \includegraphics[width=\linewidth]{fig/main/vlm_share_fm.pdf}
        \caption{$I_{\mathrm{FM}}$ by token type}
        \label{fig:a15_vlm_traj_share}
    \end{subfigure}
    \hfill
    \begin{subfigure}[t]{0.32\linewidth}
        \centering
        \includegraphics[width=\linewidth]{fig/main/vlm_share_ce.pdf}
        \caption{$I_{\mathrm{CE}}$ by token type}
        \label{fig:a15_vlm_coc_share}
    \end{subfigure}

    \vspace{0.4em}
    \begin{subfigure}[t]{0.32\linewidth}
        \centering
        \includegraphics[width=\linewidth]{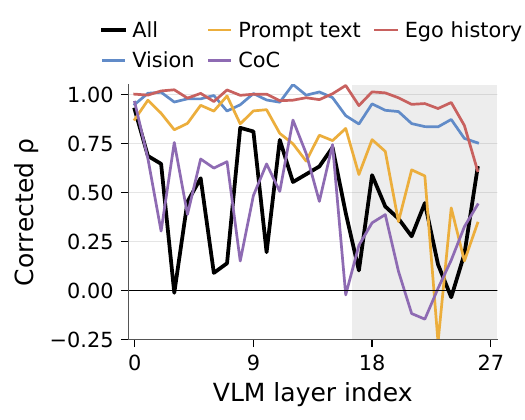}
        \caption{Rank correlation}
        \label{fig:a2e_vlm_rank}
    \end{subfigure}
    \hfill
    \begin{subfigure}[t]{0.32\linewidth}
        \centering
        \includegraphics[width=\linewidth]{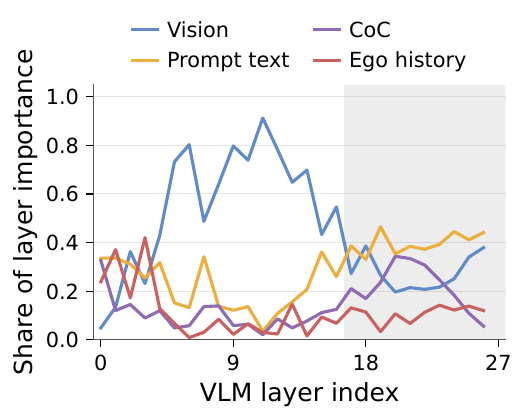}
        \caption{$I_{\mathrm{FM}}$ by token type}
        \label{fig:a2e_vlm_traj_share}
    \end{subfigure}
    \hfill
    \begin{subfigure}[t]{0.32\linewidth}
        \centering
        \includegraphics[width=\linewidth]{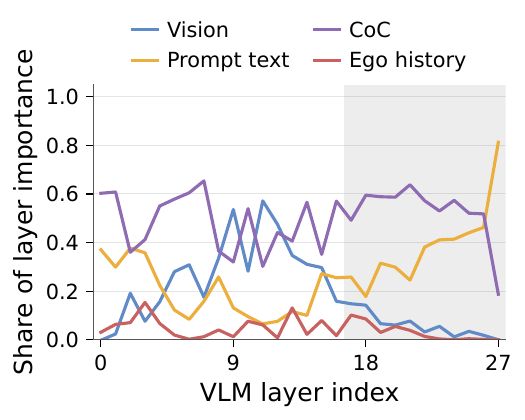}
        \caption{$I_{\mathrm{CE}}$ by token type}
        \label{fig:a2e_vlm_coc_share}
    \end{subfigure}
    \caption{\textbf{VLM analysis on Alpamayo~1.5 (a--c) and Alpamayo 2 Edge (d--f).} Panels are defined as in Figure~\ref{fig:vlm_importance}. Gray shading marks the late layers, placed at the same relative depth in both models (layers 22--35 in Alpamayo~1.5 and 17--27 in Alpamayo 2 Edge).}
    \label{fig:a2e_vlm}
\end{figure}

\paragraph{VLM.}
Unlike in Alpamayo~1.5, where the CE and FM rankings diverge mainly in the late layers, the all-token correlation of query heads in Alpamayo 2 Edge is already low in the early layers. In these layers, CE importance is concentrated on generated CoC tokens, whereas FM importance is concentrated on vision tokens (Figures~\ref{fig:a2e_vlm_rank}--\ref{fig:a2e_vlm_coc_share}). As in Alpamayo~1.5, however, the correlation computed on CoC tokens remains low across the backbone. The two objectives therefore favor different units even for the same CoC tokens, consistent with our analysis in Section~\ref{sec:vlm_analysis}.

\begin{figure}[p]
    \centering
    \begin{subfigure}[t]{0.32\linewidth}
        \centering
        \includegraphics[width=\linewidth]{fig/main/expert_rank_qhead.pdf}
        \caption{Q head rank similarity}
        \label{fig:a15_expert_q_rank}
    \end{subfigure}
    \hfill
    \begin{subfigure}[t]{0.32\linewidth}
        \centering
        \includegraphics[width=\linewidth]{fig/main/expert_rank_mlp.pdf}
        \caption{MLP channel rank similarity}
        \label{fig:a15_expert_mlp_rank}
    \end{subfigure}
    \hfill
    \begin{subfigure}[t]{0.32\linewidth}
        \centering
        \includegraphics[width=\linewidth]{fig/main/expert_cumulative.pdf}
        \caption{Cumulative importance}
        \label{fig:a15_expert_cum}
    \end{subfigure}

    \vspace{0.4em}
    \begin{subfigure}[t]{0.32\linewidth}
        \centering
        \includegraphics[width=\linewidth]{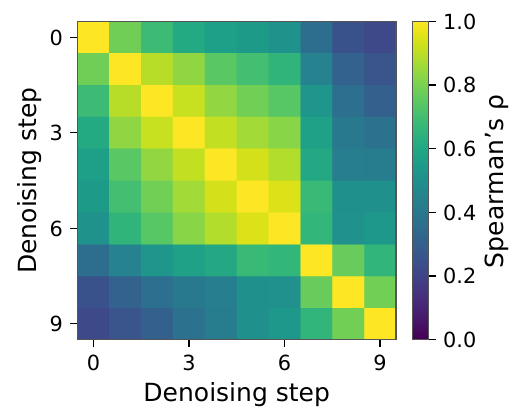}
        \caption{Q head rank similarity}
        \label{fig:a2e_expert_q_rank}
    \end{subfigure}
    \hfill
    \begin{subfigure}[t]{0.32\linewidth}
        \centering
        \includegraphics[width=\linewidth]{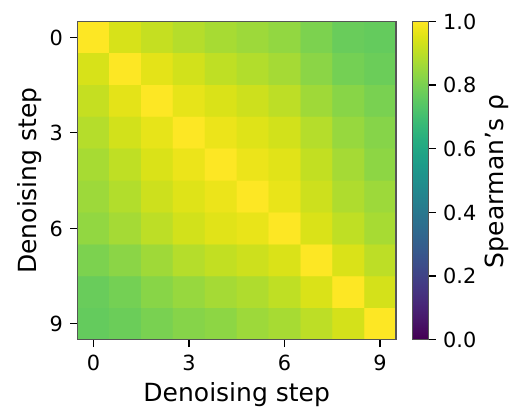}
        \caption{MLP channel rank similarity}
        \label{fig:a2e_expert_mlp_rank}
    \end{subfigure}
    \hfill
    \begin{subfigure}[t]{0.32\linewidth}
        \centering
        \includegraphics[width=\linewidth]{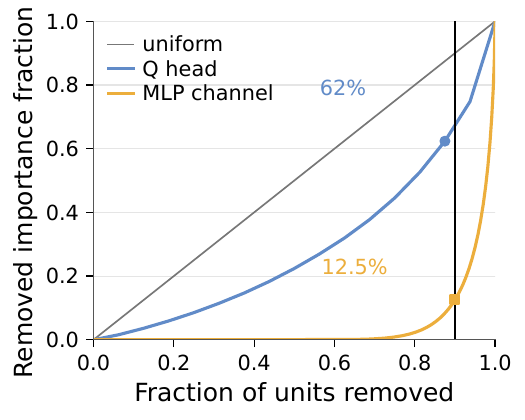}
        \caption{Cumulative importance}
        \label{fig:a2e_expert_cum}
    \end{subfigure}
    \caption{\textbf{Action expert analysis on Alpamayo~1.5 (a--c, as in Fig.~\ref{fig:expert_importance}) and Alpamayo 2 Edge (d--f).} (a, d) Pairwise Spearman correlations of FM-based Taylor importance rankings of Q heads across ten denoising steps. (b, e) The same for MLP channels. (c, f) Fraction of the measured importance removed as units are removed in ascending order of importance. The dashed diagonal represents a uniform importance distribution, and the vertical line marks 90\% removal.}
    \label{fig:a2e_expert}
\end{figure}

\paragraph{Action expert.}
MLP-channel rankings are again more stable across denoising steps than query-head rankings, with a mean Spearman $\rho$ of 0.885 versus 0.614 over step pairs (Figures~\ref{fig:a2e_expert_q_rank} and~\ref{fig:a2e_expert_mlp_rank}). MLP importance is also concentrated, although less strongly than in Alpamayo~1.5, as the lowest-scoring 90\% of MLP channels account for 12.5\% of the measured importance, compared with 1.8\% in Alpamayo~1.5.

\paragraph{Pruning.}
We also prune Alpamayo 2 Edge with \ours{}, FM-only, CE-only and the three baselines of Appendix~\ref{app:compared_methods}, ported with their original recipes (Table~\ref{tab:a2e_pruning}). \ours{}, FM-only and CE-only remove half of the query heads and MLP channels in every VLM layer, and the baselines remove approximately the same number of VLM parameters. \ours{} gives the lowest open-loop errors on all three evaluation datasets, ahead of FM-only, CE-only and the baselines, as in Alpamayo~1.5. All its margins are significant in minADE@6, except over FM-only on OOD-Val262. Expert pruning does not increase the errors.

\begin{table}[!h]
\centering
\caption{\textbf{Open-loop prediction of pruned Alpamayo 2 Edge at 6.4\,s (m).} The row with expert pruning additionally removes 75\% of the expert MLP channels. The best result is in \textbf{bold} and the second best is \underline{underlined}.}
\label{tab:a2e_pruning}
\setlength{\tabcolsep}{4pt}
\renewcommand{\arraystretch}{1.25}
\resizebox{\linewidth}{!}{%
\begin{tabular}{l|c|c|cc|cc|cc}
\toprule
\multirow{2}{*}{Method} & Expert & Pruning & \multicolumn{2}{c|}{Test500} & \multicolumn{2}{c|}{OOD-Val262} & \multicolumn{2}{c}{OOD-train} \\
 & Pruning & Ratio & minADE@6 (m)\,$\downarrow$ & minFDE@6 (m)\,$\downarrow$ & minADE@6 (m)\,$\downarrow$ & minFDE@6 (m)\,$\downarrow$ & minADE@6 (m)\,$\downarrow$ & minFDE@6 (m)\,$\downarrow$ \\
\midrule\midrule
Baseline & \xmark & 0\% & 0.916 & 2.722 & 0.964 & 2.444 & 0.975 & 2.480 \\
\midrule
Wanda & \xmark & 24.4\% & 2.407 & 6.741 & 2.523 & 6.594 & 2.666 & 6.555  \\
LLM-Pruner & \xmark & 24.4\% & 2.236 & 6.195 & 2.391 & 6.057 & 2.343 & 5.836  \\
T\'yr-the-Pruner & \xmark & 24.4\% & 1.616 & 4.466 & 1.912 & 4.775 & 2.039 & 4.946  \\
\midrule
FM-only & \xmark & 24.4\% & 1.441 & 3.881 & 1.422 & 3.652 & 1.457 & 3.601  \\
CE-only & \xmark & 24.4\% & 1.489 & 3.942 & 1.678 & 4.094 & 1.456 & 3.457  \\
\rowcolor{gray!15}
\textbf{\ours{} (Ours)} & \xmark & 24.4\% & \underline{1.196} & \underline{3.406} & \underline{1.374} & \underline{3.449} & \underline{1.322} & \underline{3.307}  \\
\rowcolor{gray!15}
\textbf{\ours{} (Ours)} & \cmark & 27.5\% & \textbf{1.190} & \textbf{3.396} & \textbf{1.363} & \textbf{3.415} & \textbf{1.318} & \textbf{3.295}  \\
\bottomrule
\end{tabular}}
\end{table}

\section{Qualitative Results}
\label{app:qualitative_results}

This section shows qualitative examples from the three evaluation protocols of the main paper: driving question answering on LingoQA (\S\ref{app:lingoqa_qual}), open-loop trajectory prediction (\S\ref{app:open_loop_qual}), and closed-loop driving in AlpaSim (\S\ref{app:closed_loop_qual}).
\S\ref{app:multi_answer} closes with a small class of multi-answer scenes, in which the closed-loop score can under-rate safe driving.

All compressed models are pruned from the same Alpamayo~1.5-10B model~\citep{nvidia2025alpamayo}, shown as Baseline.
Wanda~\citep{sun2024simple}, T\'yr-the-Pruner~\citep{li2025trthepruner}, and LLM-Pruner~\citep{ma2023llm} are the external pruning baselines.
CE-only and FM-only are single-objective variants of our criterion that score units with only the reasoning cross-entropy or only the flow-matching trajectory loss.
\ours{} is our dual-objective method.

\subsection{Driving Question Answering on LingoQA}
\label{app:lingoqa_qual}

Each panel shows four frames of the input clip, the question, the two
reference answers, and one answer per model. Each answer is scored by Lingo-Judge~\citep{marcu2024lingoqa} on a 0--100 scale, shown at the right of its row. A \textcolor{green!50!black}{$\checkmark$} and green row marks an answer the judge accepts (score above 50), and a \textcolor{red!70!black}{$\times$} and red row marks one it rejects. Answers longer than two lines are truncated
with an ellipsis. The examples are grouped in pairs by the skill they test:
signal state (Fig.~\ref{fig:lq_signal}), road users and hazards
(Fig.~\ref{fig:lq_users}), and actions at zebra crossings
(Fig.~\ref{fig:lq_zebra}).

\begin{figure}[!h]
  \centering
  \SuppTag{a}
  \SuppPanel{\linewidth}{\SuppLingoPDF}{2}{4.8 53.9 4.8 5.0}{0.43}
  \par\medskip
  \SuppTag{b}
  \SuppPanel{\linewidth}{\SuppLingoPDF}{6}{4.8 52.9 4.8 5.0}{0.43}
  \caption{\textbf{LingoQA: signal state.}
    (a)~FM-only states that there are no traffic lights.
    (b)~T\'yr-the-Pruner and CE-only make the same error.
    \ours{} reports the green state in both cases.}
  \label{fig:lq_signal}
\end{figure}

\begin{figure}[!h]
  \centering
  \SuppTag{a}
  \SuppPanel{\linewidth}{\SuppLingoPDF}{3}{4.8 52.9 4.8 5.0}{0.43}\par\medskip
  \SuppTag{b}
  \SuppPanel{\linewidth}{\SuppLingoPDF}{4}{4.8 53.9 4.8 5.0}{0.43}
  \caption{\textbf{LingoQA: road users and hazards.} (a)~LLM-Pruner answers that it is
    following a car and mentions the cyclist only as a secondary detail.
    FM-only describes the ego vehicle's own maneuver and never names the
    cyclist. (b)~The Baseline answers ``No'' even though its own
    justification mentions a pedestrian on the zebra crossing. \ours{}
    answers ``Yes'' and names the pedestrian.}
  \label{fig:lq_users}
\end{figure}

\begin{figure}[!h]
  \centering
  \SuppTag{a}
  \SuppPanel{\linewidth}{\SuppLingoPDF}{1}{4.8 43.4 4.8 5.0}{0.43}\par\medskip
  \SuppTag{b}
  \SuppPanel{\linewidth}{\SuppLingoPDF}{5}{4.8 44.4 4.8 5.0}{0.43}
  \caption{\textbf{LingoQA: actions at zebra crossings.} (a)~The Baseline
    claims that the crossing is clear and continues. \ours{} decelerates for the
    pedestrian, as in the reference. (b)~When the crossing is clear, the
    Baseline and \ours{} both continue. Every other model answers incorrectly:
    Wanda emits tokens, T\'yr-the-Pruner and LLM-Pruner return a fragment,
    CE-only answers ``Low'', and FM-only writes an incoherent sentence.}
  \label{fig:lq_zebra}
\end{figure}

\clearpage
\subsection{Open-Loop Trajectory Prediction}
\label{app:open_loop_qual}

Each panel compares the Alpamayo~1.5-10B~\citep{nvidia2025alpamayo} (Baseline) with \ours{} on one clip from OOD-Val262 (Sec.~\ref{app:open_loop_sampling}). At the top right of each panel, the best of six sampled trajectories for each
model is plotted against the ground truth over the 6.4\,s horizon. Below that,
a zoomed view covers the dashed box. The reasoning each model generated
appears beneath. A green row agrees with the reference reasoning and a red
row does not. The \emph{speed change} and \emph{cruise} labels are scenario
buckets derived only from the speed profile of the ground-truth path. A scene in which the
vehicle slows only gradually is labeled \emph{cruise}, such as creeping
behind a stopped lead vehicle (Fig.~\ref{fig:ol_follow}a) or easing off
behind a truck (Fig.~\ref{fig:ol_follow}b).

\begin{figure}[!h]
  \centering
  \SuppPanel{\linewidth}{\SuppOpenPDF}{2}{5.0 15.3 5.0 5.0}{0.43}
  \caption{\textbf{Open loop: deceleration behind a lead vehicle.} Both models
    state the same, correct intent. The difference lies in the trajectory: the
    Baseline's best sample ends about 10\,m short of the ground-truth endpoint,
    and \ours{}'s ends about 3\,m short (minADE@6 3.03\,m vs.\ 0.94\,m).}
  \label{fig:ol_lead}
\end{figure}

\begin{figure}[p]
  \centering
  \SuppTag{a}
  \SuppPanel{\linewidth}{\SuppOpenPDF}{3}{5.0 15.3 5.0 5.0}{0.43}\par\medskip
  \SuppTag{b}
  \SuppPanel{\linewidth}{\SuppOpenPDF}{5}{5.0 15.3 5.0 5.0}{0.43}
  \caption{\textbf{Open loop: following a lead vehicle.} (a)~Both models
    reason about the stopped lead vehicle, and both best samples stop short
    of the ground truth. \ours{}'s stops closer, a small gain (0.55\,m
    vs.\ 0.47\,m), and the Baseline's trajectory is largely hidden beneath
    \ours{}'s. (b)~The Baseline reasons that it must wait for a gap to merge
    left, and its best sample drifts about 1.5\,m to the left while running
    ahead of the ground truth. \ours{} reasons about keeping its distance to
    the lead vehicle, which is the truck ahead, and stays on the ground-truth
    path (3.72\,m vs.\ 0.35\,m).}
  \label{fig:ol_follow}
\end{figure}

\begin{figure}[p]
  \centering
  \SuppTag{a}
  \SuppPanel{\linewidth}{\SuppOpenPDF}{1}{5.0 15.3 5.0 5.0}{0.43}\par\medskip
  \SuppTag{b}
  \SuppPanel{\linewidth}{\SuppOpenPDF}{4}{5.0 15.3 5.0 5.0}{0.43}
  \caption{\textbf{Open loop: yielding and passing.} Both models give
    identical reasoning in both cases. \ours{}'s trajectory places the stop
    (a) and the lateral offset (b) closer to the ground truth (0.85\,m
    vs.\ 0.32\,m and 0.45\,m vs.\ 0.20\,m).}
  \label{fig:ol_yield}
\end{figure}

\clearpage
\subsection{Closed-Loop Driving}
\label{app:closed_loop_qual}

Each figure includes the Baseline and \ours{},
while other methods are shown only if they exhibit an at-fault event in the scene.
Detailed behaviors and failure cases are discussed in the individual figure captions.

\begin{figure}[!h]
    \centering
    \includegraphics[width=0.9\textwidth]{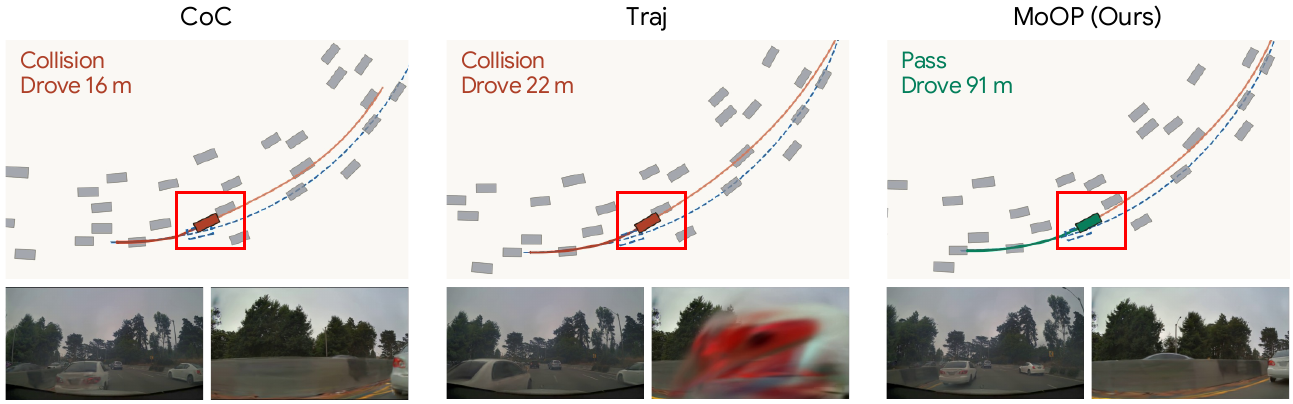}
    \caption{
    \textbf{Qualitative comparison of closed-loop evaluation.}
    Top shows the driving trajectory, while the bottom shows front and side views from the vehicle. CoC and Traj denote pruning with CE importance only and with FM importance only, respectively.}
    \label{fig:obj_comp_qual}
\end{figure}

Figure~\ref{fig:obj_comp_qual} first compares the two single-objective variants with \ours{} in one scene: CE-only and FM-only both collide, whereas \ours{} passes.

In the collision figures (all except Figs.~\ref{fig:obj_comp_qual} and~\ref{fig:cl_signal}), the top row
shows two front-camera frames from the AlpaSim~\citep{alpasim_2025}
simulation at the indicated times. Below them is one bird's-eye strip per
method, showing:
\begin{itemize}
  \item ego poses in the method's color (lighter is earlier),
  \item the plan held at the critical moment (solid),
  \item the human-driven path (dashed gray), and
  \item other agents (gray).
\end{itemize}
A $\times$ marks the ego at the method's critical moment, which is a
collision or, for T\'yr-the-Pruner in Fig.~\ref{fig:cl_wet}, leaving the
road. The red outline marks the vehicle that the method struck. In a strip
without a collision, it marks a vehicle struck by another method, drawn for
reference. Each strip's title gives the rollout's outcome, and the line below
it quotes the method's reasoning at the critical moment.
Fig.~\ref{fig:cl_signal} uses a separate layout, described in its caption.

\SuppClosedCaseNp{3}{23.8 578.3 20.7 21.5}
  {\textbf{Closed loop: cut-in.} The Baseline is hit (not at fault), and
    LLM-Pruner and CE-only have at-fault collisions. LLM-Pruner states the
    same intent as \ours{}, to keep its distance to the slow lead vehicle.
    CE-only instead proposes a lane change to the left. \ours{} keeps its
    distance and has no collision.}
  {fig:cl_cutin}

\SuppClosedCase{8}{23.8 295 20.4 21.5}
  {\textbf{Closed loop: signal changing green$\to$amber$\to$red.} Top: three
    camera frames at green, amber, and red, each with a magnified inset of the
    signal. Below: one strip per method along the approach, with the point
    where the human driver stopped (black line), the point where the method
    crossed that line (red dot), and the method's reasoning at four times.
    The Baseline, FM-only, and \ours{} stop at the line. LLM-Pruner crosses at
    16.7\,s and brakes to 0.7\,m/s, 6.8\,m past the line. T\'yr-the-Pruner
    crosses on amber at 13.4\,s at 15.1\,m/s, and Wanda and CE-only cross
    after the signal turned amber, at 17.2 and 11.7\,m/s. None of these three
    stops, and they end 64--104\,m past the line. Wanda also has an at-fault
    collision at 9.0\,s, before reaching the intersection.}
  {fig:cl_signal}

\SuppClosedCase{1}{23.8 317.6 20.7 21.5}
  {\textbf{Closed loop: bus stopped in the right part of the ego lane.}
    T\'yr-the-Pruner, Wanda, CE-only, and FM-only have at-fault collisions.
    The Baseline keeps its distance behind the bus, and \ours{} reasons that
    the bus blocks the right side of the lane and nudges left. Neither
    collides.}
  {fig:cl_bus}

\SuppClosedCase{2}{23.8 380.6 20.7 21.5}
  {\textbf{Closed loop: stopped queue ahead.} LLM-Pruner, Wanda, and
    CE-only have at-fault collisions. LLM-Pruner and CE-only state the same
    intent as the Baseline and \ours{}, to stop behind the stationary lead
    vehicle. The Baseline and \ours{} stop without contact.}
  {fig:cl_queue}

\SuppClosedCase{4}{23.8 472.1 20.7 21.5}
  {\textbf{Closed loop: slow lead vehicle.} Wanda and CE-only have
    at-fault collisions although all four methods state the same intent, to
    keep their distance to the slowing lead vehicle. The Baseline and \ours{}
    have no collision.}
  {fig:cl_slowlead}

\SuppClosedCase{6}{23.8 474.3 20.7 21.5}
  {\textbf{Closed loop: vehicle merging from the left under low sun.} Wanda
    and CE-only have at-fault collisions. CE-only proposes to nudge left
    toward a traffic island. The Baseline and \ours{} both reason about the
    car merging from the left, nudge right, and have no collision.}
  {fig:cl_lmerge}

\SuppClosedCase{7}{23.8 471.6 20.7 21.5}
  {\textbf{Closed loop: slow queue ahead.} Wanda and CE-only have at-fault
    collisions. CE-only states the same intent as the Baseline and \ours{},
    to keep its distance to the slowing lead vehicle. The Baseline and
    \ours{} keep their distance and have no collision.}
  {fig:cl_slowqueue}

\SuppClosedCase{5}{23.0 419.3 22.7 21.5}
  {\textbf{Closed loop: dense traffic on a wet road.} The Baseline is hit
    (not at fault), LLM-Pruner and Wanda have at-fault collisions, and
    T\'yr-the-Pruner leaves the road while stating the same intent as
    \ours{}. \ours{} has no collision.}
  {fig:cl_wet}

In most cases, closed-loop evaluation faithfully scores the driving performance. In \S\ref{app:multi_answer}, we discuss the exception we observed: scenes that admit more than one valid route.

\clearpage
\subsection{Multi-Answer Cases in Closed-Loop Evaluation}
\label{app:multi_answer}

As the examples in \S\ref{app:closed_loop_qual} show, the closed-loop score reflects driving quality in most scenes.
The exception we observed is the multi-answer case: a driving situation that admits more than one acceptable behavior, of which the protocol rewards only one.
At an intersection, for example, both turning and going straight can be valid.
Each closed-loop scene, however, has a single reference: the path the human driver took when the scene was recorded.
The model is given no navigation command, so nothing tells it which of the valid options that path follows.
AlpaSim~\citep{alpasim_2025} measures progress along the recorded path and stops scoring once the ego is more than $4$\,m from it.
A model that chooses a valid option other than the recorded one therefore receives a lower score than its driving warrants.

Figs.~\ref{fig:ma_turn_a}--\ref{fig:ma_turn_e} show scenes where the unpruned
Baseline scores below \ours{} for this reason. In three of them, the Baseline
states a route choice that the scene supports (a clear side street, a right-turn signal it reads as green, cross-traffic that has
cleared) and takes it, while the recorded driver goes the other way. In the
fourth, the recorded driver bears right at a fork and the Baseline keeps its
lane. Scoring stops where the two paths separate, so the Baseline is credited
with nothing it drives afterwards, whereas \ours{} follows the recorded route
and scores $1.000$ in all four. In these scenes, the lower score of the
Baseline reflects which route it chose rather than how well it drove.

To quantify how often this occurs, we count a rollout as taking an alternative route when scoring stopped under the 4\,m rule and the ego then continued for at least 10\,m without any off-road event or collision. We also estimate the score lost to the rule for every rollout stopped by it, not only for those on alternative routes. The lost score is the additional progress score that the ego accrued along the reference route after scoring stopped, counted until the end of the rollout. Table~\ref{tab:route_abs} reports both quantities for each method.

\begin{table}[!h]
\centering
\caption{\textbf{Fraction of alternative routes and score lost to the 4\,m rule.}}
\label{tab:route_abs}
\resizebox{\linewidth}{!}{%
\begin{tabular}{lcccc}
\toprule
\multirow{2}{*}{Method} & \multicolumn{2}{c}{Origin150} & \multicolumn{2}{c}{Hard100} \\
\cmidrule(lr){2-3}\cmidrule(lr){4-5} & Alt-route (\%) & Loss from 4\,m rule & Alt-route (\%) & Loss from 4\,m rule \\
\midrule
Baseline         & 6.3  & 0.037 & 10.5 & 0.063 \\
Wanda            & 5.3  & 0.040 & 9.0  & 0.094 \\
LLM-Pruner       & 5.0  & 0.022 & 14.0 & 0.067 \\
T\'yr-the-Pruner & 7.7  & 0.030 & 11.0 & 0.071 \\
FM-only          & 6.3  & 0.045 & 13.5 & 0.064 \\
CE-only          & 12.3 & 0.083 & 11.5 & 0.139 \\
\midrule
\textbf{\ours{}}          & 5.7  & 0.039 & 12.5 & 0.067 \\
\textbf{\ours{} w/ Expert Pruning} & 8.0  & 0.040 & 13.5 & 0.064 \\
\bottomrule
\end{tabular}}
\end{table}

\paragraph{Does deviation from the reference route change the method ranking?} The alternative-route fraction is similar across methods, at 5.0–8.0\% on Origin150 (except 12.3\% for CE-only) and 9.0–14.0\% on Hard100, and its differences between MoOP and Týr-the-Pruner or LLM-Pruner are not significant. The score lost to the 4\,m rule does not favor MoOP either. On Origin150, MoOP loses slightly more score to this rule (0.039 and 0.040 with expert pruning) than the Baseline, LLM-Pruner and Týr-the-Pruner (0.022–0.037), and on Hard100 the losses are similar (0.064–0.067 vs. 0.063–0.071). CE-only and FM-only lose more on Origin150 (0.083 and 0.045), but MoOP's margins over them are much larger than these differences. Even if this progress were credited back to every method, MoOP would still achieve the highest scene score among the methods at the same budget on both sets (0.867 on Origin150 and 0.663 on Hard100). The route dependence therefore lowers absolute scene scores to a similar degree across methods and does not account for the differences between them.

The ambiguity could be removed by conditioning the policy on a navigation
command or by scoring against all drivable routes consistent with the scene. We
leave this to future work.

\SuppRouteCase{1}{\textbf{Multi-answer: turning right at a residential crossroads.} The recorded driver goes straight, and the Baseline turns right after reasoning that the side street is clear. Scoring stops at the branch (2), and the rollout scores $0.221$ ($0.223$ over the scene) although it continues to drive; \ours{} goes straight and scores $1.000$.}{fig:ma_turn_a}

\SuppRouteCase{2}
  {\textbf{Multi-answer: a signalled right turn.} The recorded driver goes
    straight, and the Baseline moves into the right-turn pocket and turns after
    reasoning that the right-turn light is green. Scoring stops at the branch
    (2), and the rollout scores $0.360$ ($0.680$ over the scene); \ours{} goes
    straight and scores $1.000$.}
  {fig:ma_turn_c}

\SuppRouteCase{3}
  {\textbf{Multi-answer: turning right once cross-traffic clears.} The recorded
    driver continues straight, and the Baseline turns right after reasoning that
    cross-traffic has cleared. Scoring stops at the branch (2), and the rollout
    scores $0.560$ ($0.780$ over the scene); \ours{} continues straight and
    scores $1.000$.}
  {fig:ma_turn_d}

\SuppRouteCase{4}{\textbf{Multi-answer: a fork on a multi-lane road.} Here the recorded driver leaves the straight road and bears right, while the Baseline keeps its lane behind the lead vehicle. Scoring stops at the branch (2), and the rollout scores $0.563$ ($0.782$ over the scene); \ours{} follows the recorded fork and scores $1.000$.}{fig:ma_turn_e}

\end{document}